\documentclass[acmlarge,screen,prologue,table,xcdraw]{acmart}

\usepackage{xcolor}
\usepackage{verbatim}
\usepackage{xspace}
\usepackage{algorithm}
\usepackage{units}
\usepackage{graphicx}
\usepackage{url}
\usepackage{array}
\usepackage{multirow}
\usepackage{caption}
\usepackage{subfig}
\usepackage{algpseudocode}
\usepackage{setspace}
\usepackage{bbm}
\usepackage{makecell}
\newcommand{\tabincell}[2]{\begin{tabular}{@{}#1@{}}#2\end{tabular}}
\usepackage{booktabs}
\usepackage[normalem]{ulem}
\usepackage{aecompl}
\usepackage{dblfloatfix}
\usepackage{bm}
\usepackage{nomencl}
\makenomenclature
\usepackage{subfig}
\usepackage{float}

\specialcomment{Notes}{%
	\noindent$\bf\Rightarrow\Rightarrow$%
}{%
	$\Leftarrow\Leftarrow$%
}

\providecommand{\Note}[1]{%
	\noindent{\bf$\Rightarrow\Rightarrow$#1$\Leftarrow\Leftarrow$}%
}

\newcommand{\Donothing}{\ignorespaces\ifhmode\unskip\fi}

\newcommand{\HideNotes}{%
	\excludecomment{Notes}%
	\renewcommand{\Note}[1]{\Donothing}%
}

\HideNotes

\usepackage{xcolor}
\usepackage{microtype}      

\setcopyright{acmcopyright}
\acmJournal{IMWUT}
\acmYear{2021} \acmVolume{5} \acmNumber{2} \acmArticle{56} \acmMonth{6} \acmPrice{15.00}\acmDOI{10.1145/3463509}

\begin{document}
\title{MemX: An Attention-Aware Smart Eyewear System for Personalized Moment Auto-capture}


\author{Yuhu Chang}
\email{yhchang14@fudan.edu.cn}
\affiliation{%
  \institution{School of Computer Science, Fudan University}
  \city{Shanghai}
  \country{China}
  \postcode{200438}
}
\affiliation{%
  \institution{Shanghai Key Laboratory of Data Science, Fudan University}
  \city{Shanghai}
  \country{China}
  \postcode{200438}
}

\author{Yingying Zhao}
\authornote{Corresponding authors}
\email{yingyingzhao@fudan.edu.cn}
\affiliation{%
  \institution{School of Computer Science, Fudan University}
  \city{Shanghai}
  \country{China}
  \postcode{200438}
}
\affiliation{%
  \institution{Shanghai Key Laboratory of Data Science, Fudan University}
  \city{Shanghai}
  \country{China}
  \postcode{200438}
}

\author{Mingzhi Dong}
\email{mingzhidong@gmail.com}
\affiliation{%
  \institution{School of Computer Science, Fudan University}
  \city{Shanghai}
  \country{China}
  \postcode{200438}
}
\affiliation{%
  \institution{Shanghai Key Laboratory of Data Science, Fudan University}
  \city{Shanghai}
  \country{China}
  \postcode{200438}
}

\author{Yujiang Wang}
\email{yujiang.wang14@imperial.ac.uk}
\affiliation{%
  \institution{Department of Computing, Imperial College London}
  \city{London}
  \country{United Kingdom}
}

\author{Yutian Lu}
\email{20210240098@fudan.edu.cn}
\affiliation{%
  \institution{School of Computer Science, Fudan University}
  \city{Shanghai}
  \country{China}
  \postcode{200438}
}
\affiliation{%
  \institution{Shanghai Key Laboratory of Data Science, Fudan University}
  \city{Shanghai}
  \country{China}
  \postcode{200438}
}

\author{Qin Lv}
\email{qin.lv@colorado.edu}
\affiliation{%
  \institution{University of Colorado Boulder}
  \city{Boulder}
  \state{Colorado}
  \country{United States}
}

\author{Robert P. Dick}
\email{dickrp@umich.edu}
\affiliation{%
  \institution{University of Michigan}
  \city{Ann Arbor}
  \state{Michigan}
  \country{United States}
}

\author{Tun Lu}
\email{lutun@fudan.edu.cn}
\affiliation{%
  \institution{School of Computer Science, Fudan University}
  \city{Shanghai}
  \country{China}
  \postcode{200438}
}
\affiliation{%
  \institution{Shanghai Key Laboratory of Data Science, Fudan University}
  \city{Shanghai}
  \country{China}
  \postcode{200438}
}

\author{Ning Gu}
\email{ninggu@fudan.edu.cn}
\affiliation{%
  \institution{School of Computer Science, Fudan University}
  \city{Shanghai}
  \country{China}
  \postcode{200438}
}
\affiliation{%
  \institution{Shanghai Key Laboratory of Data Science, Fudan University}
  \city{Shanghai}
  \country{China}
  \postcode{200438}
}

\author{Li Shang}
\authornotemark[1]
\email{lishang@fudan.edu.cn}
\affiliation{%
  \institution{School of Computer Science, Fudan University}
  \city{Shanghai}
  \country{China}
  \postcode{200438}
}
\affiliation{%
  \institution{Shanghai Key Laboratory of Data Science, Fudan University}
  \city{Shanghai}
  \country{China}
  \postcode{200438}
}

\renewcommand{\shortauthors}{Chang et al.}

\begin{abstract}
This work presents \emph{MemX}: a biologically-inspired attention-aware eyewear system developed with the goal of pursuing the long-awaited vision of a personalized visual Memex. \emph{MemX} captures human visual attention on the fly, analyzes the salient visual content, and records moments of personal interest in the form of compact video snippets. Accurate attentive scene detection and analysis on resource-constrained platforms is challenging because these tasks are computation and energy intensive. We propose a new temporal visual attention network that unifies human visual attention tracking and salient visual content analysis. Attention tracking focuses computation-intensive video \textcolor{black}{analysis} on salient regions, while video \textcolor{black}{analysis} makes human attention detection and tracking more accurate. Using the YouTube-VIS dataset and 30 participants, we experimentally show that \emph{MemX} significantly improves the attention tracking accuracy over the \textcolor{black}{eye-tracking-alone} method, while maintaining high system energy efficiency. We have also conducted 11 in-field pilot studies across a range of daily usage scenarios, which demonstrate the feasibility and potential benefits of \emph{MemX}.
\end{abstract}
%
%


\maketitle

\Note{*** RD: Although it is appropriate to italicize words from Latin, it isn't appropriate to italicize et al., i.e., and e.g., as they are now considered part of the English language. You can save the italics for introducing new terms and genuine foreign phrases, e.g., \emph{Corruptissima re publica plurimae leges}.*done by YZ*}

\section{Introduction}
\label{sctn::intro}

First presented by Vannevar Bush then restated by Jim Grey, Personal Memex paints the vision of the capability to capture and record personal experiences in daily life~\cite{bush1945we}. Rapid improvement in mobile camera and signal processing technology has put us on the verge of realizing this vision. In particular, cameras in wearable forms, such as GoPro, Google Clips, and Spectacles from Snapchat, enable people to take photos and record videos anywhere, anytime. 

However, challenges remain. Personalization is the core value of Personal Memex. Videos and photos captured by devices must match people's personal interests to remain relevant and valuable. A fundamental limitation of existing wearable cameras is the inability to automatically detect and selectively record the visual content of personal interest. Most existing wearable cameras, such as GoPro and Spectacles, require manual intervention for content recording. Empowered by machine vision technologies, recently released Google Clips performs auto-capture if interesting scenes are detected. However, scene interest is estimated solely by the built-in machine vision algorithm: instead of being personalized to the user, it provides a one-size-fits-all solution. As noted by a reputable 3rd-party reviewer, quoted here ``Clips' standout feature -- artificially intelligent auto-capture -- is too unpredictable, keeping it from being truly satisfying''~\cite{Low18review}. \Note{***RD: Can you use a direct quote from the review and put in quotes?*** ***LS: Done***}

\begin{figure}[h]
	\includegraphics[width=1.0 \textwidth]{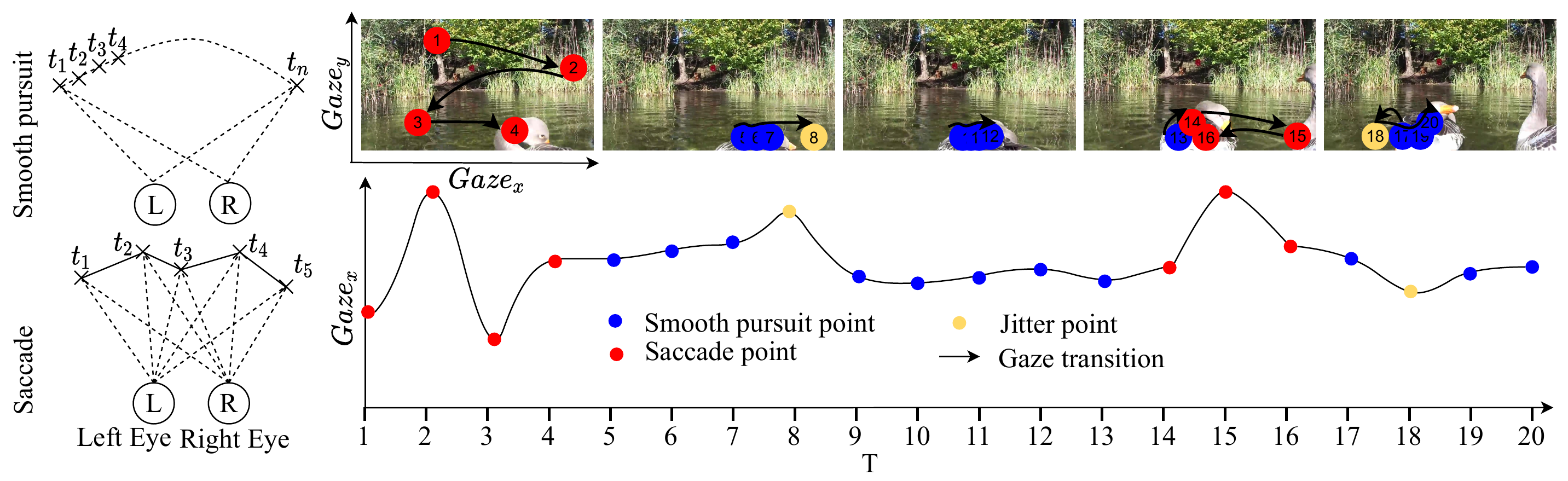}
 \caption{(Left column) shows two eye movement types: saccade (top) and smooth pursuit (down). (Right column) An example of human gaze trajectory profile, which is a dynamic and noisy process. \textcolor{black}{$Gaze_x$ and $Gaze_y$ denote the gaze position along the x axis and y axis, respectively.}}
 	\label{fig::eye_move}
 \end{figure}

Recent advances in eye-tracking technologies offer the capability to measure human visual attention~\cite{9157393,anton2013attentional}. Based on physiology and psychology research~\cite{Broadbent1957A}, human attention serves as an information filter and prioritization strategy~\cite{mancas2016human,Broadbent1957A}, which selectively determines the allocation of cognitive resources. It operates as a gate that connects human inner consciousness with the outer world~\cite{2006The}. Via attention gate, human selects the prioritized information regarding the most interesting areas or objects from a huge amount of unstructured perceived information~\cite{mancas2016human}. Physiologically, humans direct their gaze to objects of interest~\cite{nelson2018}. Human eye movement consists of several dynamic patterns, including saccade, smooth pursuit, vergence, and fixation~\cite{wang2019neuro}; saccade and smooth pursuit account for most eye movements. 
Saccades are conjugate eye motions, where gaze is rapidly shifted from one location or object to another. They can be voluntarily directed or involuntarily triggered by visual stimuli. Smooth pursuit is smooth eye motion allowing the gaze to continuously follow an object of interest. Saccades and smooth pursuit may alternate~\cite{behera2005recurrent}. They can be measured and used to produce visual attention time series, as shown in Fig.~\ref{fig::eye_move}. Eyewear systems may include, or be retrofitted with, eye-tracking technologies, e.g., inward-facing eye cameras, to measure human regional gaze fixation within the field of view, and infer the focused region of human visual attention.

Despite its promising potential, \textcolor{black}{eye tracking} suffers from limitations that complicate practical use. First, \textcolor{black}{eye tracking} does not directly detect the object of human visual interest. Instead, it estimates human visual regional focus by measuring the angular direction of eye gaze relative to the field of view of the eyewear. Measurement noise, e.g., angular offset and head motion, can easily produce incorrect visual attention measurements. In addition, eye movement is noisy. Human gaze frequently experiences transient jitter: gaze may drift away from the object of interest due to other visual stimuli or noise (As shown in Fig.~\ref{fig::eye_move}). Therefore, \textcolor{black}{eye tracking} alone is inadequate for reliable attention tracking. A potential remedy to this problem is to augment \textcolor{black}{eye tracking} with visual scene analysis using video analytics techniques, e.g., identifying the salient object of human visual focus using instance segmentation~\cite{yang2019video} and analyzing the spatial visual context using object detection and recognition~\cite{937604}. This motivates us to combine eye-tracking with video analytics to enable accurate human visual attention detection and tracking. However, machine vision techniques are computationally demanding and energy-intensive, limiting adoption in resource-constrained wearable settings. These challenges must be addressed in order to accomplish our goal of accurate and efficient human visual attention tracking.

This paper presents~\emph{MemX}, a biologically-inspired attention-aware eyewear system to enable automated, highly personalized capture of interesting visual content, which is recorded in compact video snippets. We retrofit off-the-shelf glasses with an inward-facing eye camera and a forward-facing world camera. The eye camera performs \textcolor{black}{eye tracking} to determine human gaze direction. The world camera samples and analyzes the field of view. We propose a new temporal visual attention (TVA) network, which unifies \textcolor{black}{eye tracking and video analysis}, to enable accurate and computationally efficient human visual attention tracking and salient visual content analysis. Specifically, the eye camera performs \textcolor{black}{eye tracking} to determine human visual regional focus, which confines computation-intensive video analytics tasks solely to regions of personal interest, thereby minimizing computing time and energy costs. Working in tandem, the video analytics component analyzes the visual scene content and identifies the salient object of human visual focus as well as the spatial visual context, thereby improving the accuracy and stability of human visual attention tracking. In addition, the high-resolution scene frame is used as \textcolor{black}{supplementary} input to the TVA network for more accurate attention detection. Such high-resolution scene frames are captured only when likely attention is detected, which significantly reduces energy consumption by decreasing the average sampling frequency. Furthermore, eye movement types, i.e., saccades or smooth pursuit, act as an ``intermediate'' supervision to be incorporated into the learning process. \Note{*** RD: The previous sentence isn't clear to me. LS: done *** This helps to regularize the learning problem of attention tracking, leading to more accurate decisions. *** RD: The previous paragraph is redundant about 2/3 of the way in. I can revise that tomorrow. ***}

This work makes the following contributions.
 \begin{enumerate}

\item We propose a new temporal visual attention (TVA) network that unifies eye-tracking and video analytics, to enable accurate and computation-efficient human visual attention tracking.

\item  We develop \emph{MemX}, a new biologically inspired, attention-aware eyewear system. Equipped with the proposed TVA network, \emph{MemX} automatically detects human visual attention and uses it to analyze and record visual content of personal interest.

\item \emph{MemX} is evaluated using the YouTube-VIS dataset~\cite{yang2019video} and 30 participants. Experimental results show that compared to the \textcolor{black}{eye-tracking-alone} method, \emph{MemX} significantly improves the attention tracking accuracy, while maintaining high system energy efficiency. 

\item \emph{MemX} can be applied in several personal visual content capture scenarios, such as sightseeing, lifelogging, academic events (e.g., meetings), shopping, and sports logging. We conduct 11 in-field pilot studies with different potential usage scenarios to evaluate \emph{MemX} and illustrate some of its potential uses and benefits. According to the questionnaire survey, 96.05\% of the moments of interest are successfully captured by \emph{MemX}. In addition, the pilot studies demonstrate significant energy savings as compared with the record-everything case, with 86.36\% energy savings on average.

\end{enumerate}

The rest of the paper is organized as follows. Section~\ref{sctn::related} surveys related work. Section~\ref{sctn::method} presents the proposed \emph{MemX} system and algorithm. Section~\ref{sctn::sys} details~\emph{MemX} system design and implementation. Section~\ref{sctn::exp} presents and discusses the experimental results. Section~\ref{sctn::app} presents the in-field pilot studies. We conclude the work in Section~\ref{sctn::cnclusn}.

\section{Related Work}
\label{sctn::related}

This section summarizes the most relevant work in the areas of \textcolor{black}{eye tracking}, eye movement classification, visual attention, human interest, and computation-efficient video analytics. 

\subsection{Eye Tracking}

There has been extensive work on eye tracking and estimation. Two comprehensive reviews of existing approaches are provided by Hansen~\textcolor{black}{et al.}~\cite{hansen2009eye} and Cazzato~\textcolor{black}{et al.}~\cite{cazzato2020look}.  These existing methods can be categorized into two classes: model-based approaches~\cite{wood2014eyetab, wang2015atypical} and appearance-based approaches~\cite{ zhang2017mpiigaze, zhang18_etra}. For model-based methods, high-quality images and appropriate lighting conditions are critical, while appearance-based methods leverage more facial features, such as head orientation, from large-scale training data. Compared with model-based methods, appearance-based methods are less sensitive to image quality. Recent work has shown that appearance-based methods can offer good generalization capability\Note{*** RD: To ``what'' these faces? To distinguish them or something else?'' **done by YZ: remove : to new human face*} even without user-specific data~\cite{zhang2019evaluation}. Thanks to the emerging large-scale data sets and deep learning techniques, the performance of learning-based eye-tracking models has been steadily improving~\cite{krafka2016eye, wong2019gaze, wang2019generalizing}.
\textcolor{black}{Our system unifies eye tracking and video analysis to track human visual attention, allowing video analysis to focus on salient visual content, thereby reducing computation and energy costs.}
 \Note{*** QL: How does our eye-tracking method differ from prior work? **done by YZ* }

\subsection{Eye Movement Classification}

As a fundamental step towards eye tracking~\cite{komogortsev2010qualitative}, eye movement classification has been extensively studied and a wide range of methods \textcolor{black}{have} been proposed, including classical algorithms~\cite{salvucci2000identifying}, Bayesian Mixed Models (BMM)~\cite{Bayesian2012eyemovement}, and deep learning~\cite{Raimondas2018gazeNet}. Most classical approaches identify saccade and fixation movements using velocity-threshold identification, dispersion-threshold identification, and hidden Markov model identification~\cite{djanian2019eye}. Later, Tafaj~\textcolor{black}{et al.}~\cite{Tafaj2013} and Santini~\textcolor{black}{et al.}~\cite{Santini2016Bayesian}  added eye movement classification of smooth pursuit in BMM. Recently, deep learning has been used to classify \textcolor{black}{eye movements}. For example, Zemblys~\textcolor{black}{et al.} proposed a CNN-based architecture for sequence-to-sequence eye movement classification~\cite{Raimondas2018gazeNet}. Startsev~\textcolor{black}{et al.} combined \Note{***RD: Did they propose to do it or do it? If they did it, ``combined'' is better than ``proposed to combine''. **done by YZ*} 1D-CNN with BLSTM to classify eye movements as fixation, saccade, and smooth pursuit in windows of up to \unit[1]{s}~\cite{startsev20191d}. Our system uses eye movement \textcolor{black}{types} as a supplement ``intermediate'' supervision \Note{***RD: The previous half-sentence is ambiguous. I'm not sure what you mean.**done by YZ*} to drive system-level learning in \emph{MemX}, and also identifies the instants of \textcolor{black}{visual} attention.

\subsection{Visual Attention and Eye Tracking}

Due to the limited processing capabilities of biological vision \textcolor{black}{system}, attention (and high spatial resolution capture by the fovea) is directed to a small portion of a scene~\cite{carrasco2011visual}, thereby reducing the computational demands on the visual cortex. Leveraging attention, humans can flexibly control limited computational resources~\cite{lindsay2020attention} to process the most relevant and preferential sensory information~\cite{anton2013attentional}.
Recently, extensive studies have combined visual attention and eye-tracking technologies. For instance, Hwang~\textcolor{black}{et al.} proposed to use eye-tracking to quantify human visual attention in an online shopping environment~\cite{hwang2018using}. Katz~\textcolor{black}{et al.} used eye-tracking to identify regional gaze fixation on a respiratory function monitor for medical diagnosis~\cite{katz2019visual}. Wang~\textcolor{black}{et al.} applied visual attention in unsupervised video object segmentation~\cite{wang2019learning}. 
These works promote the idea of combining eye-tracking and video analytics methods to enable accurate and efficient human attention \textcolor{black}{detection and} tracking. 
\Note{***RD: Keep the same style for each. One sentence at end to contrast your work, e.g., ``Our system also combines these sensing and analysis modes, but in contrast with prior work uses them to ***.''}

\subsection{Visual Attention and Human Interest}

Previous works in the field of physiology and psychology research demonstrate that eye movements reveal human attention and gaze positions indicate interesting and important viewed content.
Jain~\textcolor{black}{et al.} pointed out that eye movements can reveal attention's change or shift~\cite{jain2015gaze}. For example, saccade means quick visual attention shift, fixation means the visual attention rests at the same location, and smooth pursuit suggests visual attention follows a moving target smoothly~\cite{jain2015gaze}. 
Tag~\textcolor{black}{et al.}~\cite{10.1145/3027063.3053243} described a smart eyewear platform capable of attention tracking by measuring the cognitive engagement of people in different situations. Later, Abdelrahman~\textcolor{black}{et al.}~\cite{abdelrahman2019classifying} pointed out that eye data can be indicative for human cognition~\cite{abdelrahman2019classifying,2018Watch}. For instance, gaze positions can reveal the attention locus~\cite{abdelrahman2019classifying}. Similarly, Kranthi~\textcolor{black}{et al.}~\cite{2018Watch} suggested that gaze can be used as an indicator of importance when humans watch a video. That is, gaze positions indicate important or interesting viewed content~\cite{2018Watch}. 

\subsection{Computation-Efficient Machine Vision}

Deep learning has achieved great success in machine vision applications, such as object detection, object classification, and semantic segmentation. Deep models, such as ResNet~\cite{he2016deep}, DenseNet~\cite{huang2017densely}, MobileNetV2~\cite{sandler2018mobilenetv2}, and MobileNetV3~\cite{howard2019searching} are widely used in machine vision applications. However, deep models incur high computation \textcolor{black}{costs}, limiting adoption in resource-constrained mobile environments.  
\textcolor{black}{Light-weight computation-efficient network architecture is therefore an important research branch of deep learning communities.} Take for example the architecture of Google's MobileNetV2~\cite{sandler2018mobilenetv2} which is a representative light-weight network designed for mobile device. Linear bottlenecks and inverted residual blocks were used as the basic structure in MobileNetV2, aiming to solve the issue of feature degradation during training. Also, as pointed out by Lubana~\textcolor{black}{et al.}, energy consumption significantly depends on the transferred resolutions in imaging systems~\cite{lubanaDigitalFoveationEnergyAware2018}. Therefore, they proposed that energy consumption can be dramatically reduced if only the task-related information is input to deep models. Inspired by the above work, we propose a new lightweight temporal visual attention network, which fuses two visual input sources, i.e., temporal eye movements and field-of-view scene video, to offer higher computation efficiency while still guarantee task performance. \Note{*** RD: You use ``Lightweight deep models'' without definition, and I cannot see how the implied definition differs in any way from ``efficient deep models''. Therefore, ``Recent work on computation-efficient deep models focuses on designing lightweight deep models'' looks like ``Recent work on computation-efficient deep models focuses on designing efficient deep models'' to me. I think most work on efficient deep models reduce computation via operand width reduction (narrow weights); parameter elimination (structured and unstructured pruning), and input data elimination (sparse sampling).'' Then you could give a cited example of each category.**done by YW*}

\section{Temporal Visual Attention Network}
\label{sctn::method}

\subsection{Design Motivations}
The vision of the personalized Memex builds on the key observation that content capture must match a user's personal interest 
in order to be relevant and valuable. As visual attention directly reflects human attentive \Note{***RD: suggest deleting two previous words.} interest, the goal of this work is 
to design an accurate and efficient method for human visual attention tracking and salient content analysis.
In this section, we first describe human visual attention, and then discuss the capabilities and limitations
of eye-tracking and video \textcolor{black}{analysis} techniques, which motivate the proposed temporal visual analytical network, a  
unification of these two techniques.

Humans direct their foveas, the high-resolution regions of their retinas, toward locations attracting their attention: foveal orientations reveal locations of
visual attention. Since real-world scenes often contain multiple stationary and moving objects, attention is a temporally and spatially selective process, driven by top-down influences and bottom-up sensing
mechanisms. It starts with steering the eye gaze towards an object attracting attention, then continuously 
follows the object of interest, until new visual stimuli or top-down goals cause the fovea to drift away.

From the eye-tracking perspective, as stated in Section~\ref{sctn::intro} and illustrated in Fig.~\ref{fig::eye_move},
human eye movement patterns can be mainly classified into two distinct and alternating phases, namely
saccade and smooth pursuit~\cite{leigh2015neurology}. Saccade, the rapid conjugate eye shift, is triggered either voluntarily
or by an external visual stimulus, indicating the change or a new moment of human visual attention. \Note{***RD: Not quite sure what the previous sentance means.} Smooth pursuit,
on the other hand, refers to slow and smooth eye movements following the same
object of interest.  Together, a human visual attention epoch starts with a saccade phase, followed by a smooth
pursuit phase. Therefore, eye movement analysis, e.g., saccade-smooth pursuit transition detection, 
can be used to detect and classify the moments of human 
visual attention. In addition, as described in Section~\ref{sctn::related}, recent work on eye tracking provides 
relatively accurate estimation of human visual regional focus by measuring the angular direction of eye gaze relative 
to the eyewear's field of view. Furthermore, eye tracking is computationally efficient, as it requires low-resolution eye images and light-weight algorithms. However, as stated in Section~\ref{sctn::intro}, 
eye tracking does not directly detect the object of interest, and cannot independently provide accurate and stable indications of human visual interest, due to limited tracking resolution and inherently noisy eye 
movement patterns (as shown in Fig.~\ref{fig::eye_move}). \Note{***RD: Is the tracking resolution really so low it causes a problem?}

A key motivation of this work is that video \textcolor{black}{analysis} techniques, e.g., object detection and classification, offer
object or semantic level information, which is complementary to regional-level eye tracking. It is thus possible to 
augment eye tracking with video \textcolor{black}{analysis} techniques to perform accurate and stable human attention detection. 
However, object and semantic level video \textcolor{black}{analysis} tasks are data- and computation-intensive. Consider recent
work on video instance segmentation (VIS)~\cite{yang2019video}; VIS is one of the most \textcolor{black}{broadly-used video analysis} 
tasks, consisting of \textcolor{black}{simultaneous detection, segmentation, and tracking of object instances.} The latest work 
on VIS adopts ResNet-50-FPN or ResNet-100-FPN~\cite{johnson2018adapting} as a feature extractor and uses regional proposal 
network~\cite{ren2015faster} to identify possible regions of objects. However, the high energy demand of the VIS task limits its adoption in computation and energy-constrained 
wearable settings.
 
We observe that eye tracking can help improve the efficiency of video \textcolor{black}{analysis} tasks. 
First, eye tracking efficiently captures human visual regional focus, focusing video \textcolor{black}{analysis} on 
regions of interest and eliminating analysis of other, superfluous regions, thereby improving computational efficiency. 
Second, eye tracking can detect human visual attention, enabling video \textcolor{black}{analysis} to be triggered only when
human attention is likely. 
Furthermore, eye movement types, i.e., saccades or smooth pursuit, can serve as an ``intermediate'' supervision that can be 
incorporated into the learning process to regularize the learning problem, leading to more accurate decisions. \Note{***RD: The previous sentence is vague and I don't know enough about your intention to revise it.}

In summary, working in tandem, eye tracking and video \textcolor{black}{analysis} can potentially deliver accurate and efficient 
human visual attention tracking and serve as the foundation to support personalized moment auto-capture. 

\subsection{Pipeline of the TVA Network}
Here, we describe the overall pipeline of the proposed temporal visual attention (TVA) network that we deploy in the \emph{MemX} system to detect the user's visual attention.

\begin{figure}[h]
	\includegraphics[width=1.0 \textwidth]{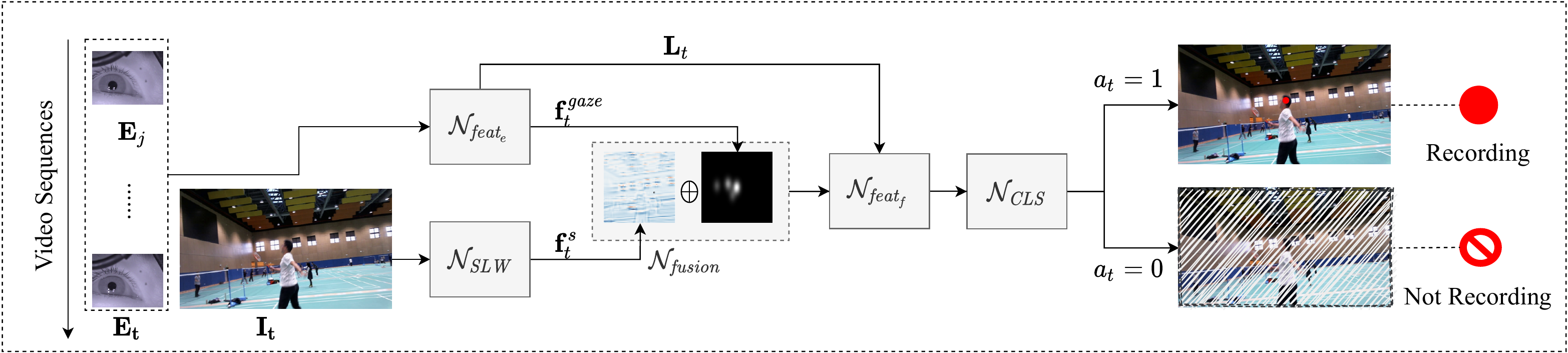}
 \caption{Overall pipeline of the proposed Temporal Visual Attention (TVA) network. The TVA network distills historical eye movement phases likelihood $\mathbf{L}_t$ and gaze features $\mathbf{f}_t^{\mathit{gaze}}$ from the low-resolution eye camera, along with object and semantic features $\mathbf{f}_t^s$ from the high-resolution world camera, and then performs feature fusion and classification for human attention detection and content auto-capture.}
 	\label{fig::system}
 \end{figure}

Fig.~\ref{fig::system} shows the proposed TVA framework that unifies eye-tracking and video \textcolor{black}{analysis} to detect the user's visual attention.
Two video streams serve as the inputs of this framework, which include (1) the scene stream
$\mathbf{I}=\{\mathbf{I}_1, \mathbf{I}_2, \dots, \mathbf{I}_m\}$ captured by the \textcolor{black}{forward-facing} world camera; 
and (2) the video sequence $\mathbf{E}=\{\mathbf{E}_1, \mathbf{E}_2, \dots, \mathbf{E}_m\}$ captured by the \textcolor{black}{inward-facing} eye camera, where $m$ is the total number of time steps.  
For a given time step $t$, the TVA network predicts whether the user is attentive to a newly-appeared instance or tracks a previously-attentive instance for the current frame $\mathbf{I}_t$ with a binary result $a_t \in \{0,1\}$.
If true ($a_t = 1$), \textcolor{black}{\emph{MemX} invokes high-resolution video recording.}
Otherwise ($a_t = 0$), the current frame $\mathbf{I}_t$ is discarded to save energy.

This work develops \emph{MemX} to enable automated, personalized capture of interesting visual content, with the energy-constrained wearable form factor. To this end, the proposed TVA network aims to deliver accurate and energy-efficient visual attention detection by unifying eye-tracking and video analysis. To generate temporally-consistent predictions, the TVA network continuously senses the inward-facing low-resolution \textcolor{black}{eye video} and extracts two essential gaze representations, which are the likelihood $\mathbf{L}_t$ of historical eye movement types and the gaze features $\mathbf{f}_t^{\mathit{gaze}}$ extracted from the historical gaze positions, as shown in Fig.~\ref{fig::system}. Those 
two gaze representations are fused together to make an initial prediction that \textcolor{black}{a potential attention} epoch is occurring.
In particular, the likelihood $\mathbf{L}_t$ of historical eye movement types are used to capture gaze saccade-smooth pursuit transition, which is a clear indicator of the beginning of an attention epoch. However,  the gaze transition likelihood $\mathbf{L}_t$ alone may not be reliable since the eye movements are subtle and difficult to identify.   
Therefore, we also use the historical gaze positions $\mathbf{P}_t$ to improve detection robustness. Intuitively, if the majority of the recent gaze positions fall into a small circular region, we can predict with higher confidence that the user's eye movement has indeed entered a smooth pursuit phase. 
Moreover, as pointed out by prior works such as \textcolor{black}{pupil detection and tracking~\cite{kassner2014pupil}, which is suitable for energy-constrained wearable scenarios, eye tracking} is computationally efficient. Eye tracking may therefore be used for every frame, which can effectively minimize the invocation of
\textcolor{black}{computation-intensive video analysis processing and energy-intensive high-resolution video recording.}

As mentioned earlier, eye tracking alone is not sufficiently accurate for attention detection. 
To boost the accuracy of attention detection, the forward-facing high-resolution world camera is also used \textcolor{black}{only} when human attention is likely.
The TVA network then extracts low-level scene features $\mathbf{f}_t^{s}$ from the high-resolution frame $\mathbf{I}_t$, and fuse it with the gaze features $\mathbf{f}_t^{\mathit{gaze}}$ extracted from historical gaze positions $\mathbf{P}_t$. Compared with the size of the original scene image (224$\times$224), the spatial resolution of the gaze features $\mathbf{f}_t^{\mathit{gaze}}$ is more compact (56$\times$56), and it can also guide our TVA network to focus on analyzing the attentive local region, which further reduces the computational cost.

The obtained scene features $\mathbf{f}_t^{s}$ are then fused with
the likelihood $\mathbf{L}_t$ of historical eye movement types and gaze features $\mathbf{f}_t^{\mathit{gaze}}$ in the TVA network to better predict the temporal attention of the user. 

Instead of treating eye-tracking and video analysis as two separate tasks, we design a light-weight network to perform feature fusion. Specifically, we first apply a shallow network $\mathcal{N}_{\mathit{SLW}}$ to obtain features $\mathbf{f}_t^s$ for frame $\mathbf{I}_t$, i.e., $\mathbf{f}_t^s=\mathcal{N}_{\mathit{SLW}}(\mathbf{I}_t)$. 
We then map the historical eye tracking results to gaze position related features $\mathbf{f}_t^{\mathit{gaze}}$ and the likelihood of eye movement phases
$\mathbf{L}_t$. The mapping function is denoted as $\mathcal{N}_{\mathit{feat}_e}$, i.e., 
$\mathbf{L}_t, \mathbf{f}_t^{\mathit{gaze}} = \mathcal{N}_{\mathit{feat}_e}(\mathbf{E}_j,\cdots, \mathbf{E}_t)$. After that, 
we fuse $\mathbf{f}_t^{\mathit{gaze}}$  and $\mathbf{f}_t^{s}$ using a predefined operation $\mathcal{N}_{\mathit{fusion}}$ that makes 
those features complementary to each other so as to augment the detectability of salient region focus. 
In addition, $\mathbf{L}_t$ is leveraged as an ``intermediate'' supervision to drive the learning process. 
The intuition is that if attention is drawn to an object, historical eye movements may follow 
a detectable pattern, which can be used as prior knowledge to supervise the learning network. For example, an eye-movement sequence: \emph{saccade}, \emph{smooth pursuit}, 
$\cdots$, \emph{smooth pursuit}, can suggest the occurrence of attention with high confidence. Finally, the above output goes through a classification model $\mathcal{N}_{\mathit{CLS}}$ to generate the final decision of $a_t$ (attention or not). 
Next, we provide the detailed implementation of the proposed TVA network.

Given an incoming eye frame $\mathbf{E}_t$, we want to detect user attention, i.e., whether the user is potentially gazing at an object within the field of view. We experimentally define the occurrence of potential attention as the majority of the gaze positions falling within a region with area $A$ or smaller during a time period $T$. The impacts of $A$ and $T$ on attention detection performance are evaluated in Section~\ref{sctn::exp}.

If possible attention is detected at time $t$, we then trigger the world camera and sense scene frame $\mathbf{I}_t$ to further boost the confidence level that the user is indeed paying attention to an object. Specifically, let $\mathbf{P}_t=\{\mathbf{p}_{t-N+1},...,\mathbf{p}_{t-1}, \mathbf{p}_{t}\}$ be the predicted historical gaze positions during time steps $(t-N,t]$ where $\mathbf{p}_k \in \mathbb{Z}^2$, and let $\mathbf{L}_t=\{\mathbf{l}_{t-N+1},...,\mathbf{l}_{t-1}, \mathbf{l}_{t}\}$ be the predicted historical eye movement types likelihood where each $\mathbf{l}_i  \in \mathbb{R} \cap [0,1]$. 
The goal of the TVA network is to determine from $\mathbf{I}_t$, $\mathbf{P}_t$ and $\mathbf{L}_t$ whether the user is actually gazing at an object within the scene frame, i.e.,  $a_t=\mathcal{N}_{TVA}(\mathbf{P}_t,\mathbf{I}_t, \mathbf{L}_t)$, and $\mathcal{N}_{\mathit{TVA}}$ represents the proposed TVA network. 

After that, a shallow network $\mathcal{N}_{SLW}$ is applied to the scene image $\mathbf{I}_t$ to generate the scene features $\mathbf{f}_t^{s}$ with size $56\times 56 \times 24$, i.e.,  $\mathbf{f}_t^{s}=\mathcal{N}_{\mathit{SLW}}(\mathbf{I}_t)$
Since the historical gaze positions $\mathbf{P}_t$ are discrete and not differentiable, we relax each historical gaze position into a continuous heatmap of size $56 \times 56$ based on Gaussian distributions. Specifically, for the $k^{th}$ gaze position $\mathbf{p}_k \in \mathbb{Z}^2$, the value $h_{\mathbf{p}_i}$ for a location $\mathbf{p}_i \in \mathbb{Z}^2$ in the heatmap $\mathbf{H}_k$ can be computed as 
\begin{equation}
\label{eq:gaussian_relax}
h_{\mathbf{p}_i} = \frac{w_k}{{\sigma \sqrt {2\pi } }}exp(-\frac{D^2(\mathbf{p}_i,\mathbf{p}_k)}{2\sigma ^2}),
\end{equation}
where $D(\mathbf{p}_i,\mathbf{p}_t)$ represents the Euclidean distance between $\mathbf{p}_i$ and $\mathbf{p}_t$, and $w_k$ is a pre-defined weight related to the time step $k$. 
Applying the Gaussian Relaxation in Eq. \ref{eq:gaussian_relax} to each gaze position in $\mathbf{P}_t$ will generate $N$ heatmaps $\{\mathbf{H}_{t-N+1},...,\mathbf{H}_{t-1}, \mathbf{H}_{t}\}$ of resolution $56 \times 56$. We denote those heatmpas as $\mathbf{f}_t^{\mathit{gaze}}$. 

Finally, The gaze heatmaps $\mathbf{f}_t^{\mathit{gaze}}$, the scene features $\mathbf{f}_t^{s}$ and their element-wise product $\mathbf{f}_t^{\mathit{gaze}} \circ \mathbf{f}_t^{s}$ 
are concatenated channel-wise to fuse information from gaze analysis and scene images. The resulting features, which are of size $56 \times 56 \times (24+8)$, 
are then fed into \textcolor{black}{one} convolution layer Conv0 with 56 output channels. \Note{***RD: Previous sentence unclear. Are there two or one layers? Did you omit ``Conv1''?} A following fully-connected layer FC0 gradually reduces the channels into two representing the probability of the binary action $a_t$. In addition, inspired by the work of Wang et al.~\cite{wang2020dynamic}, we append the likelihood $\mathbf{L}_t$ of historical eye movement types with the input tensor of FC0 layer to supervise the learning during training and help with the inference during testing.

\section{\emph{MemX}: An Attention-Aware Eyewear System}
\label{sctn::sys}
This section describes~\emph{MemX}'s hardware design and TVA network integration to support the operation 
of personalized moment auto-capture. Since energy efficiency is a key focus of \emph{MemX}, this section also 
describes \textcolor{black}{the energy model of \emph{MemX}}.

\subsection{\emph{MemX} Hardware Design}
\begin{figure}[htb]
        \includegraphics[width=1.0 \textwidth]{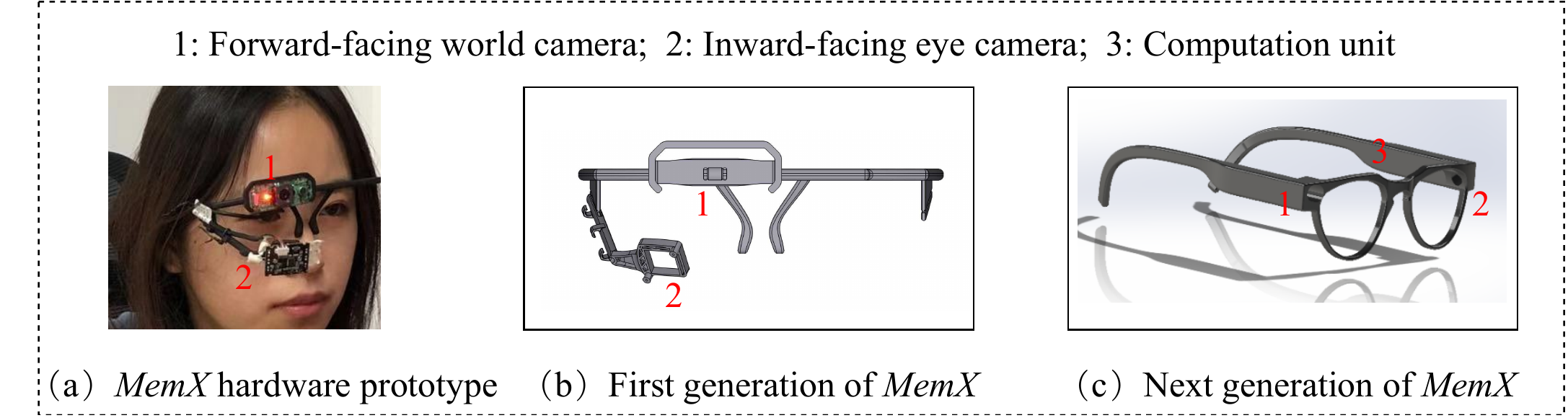}
        \caption{\emph{MemX} smart eyewear hardware prototype.}
        \label{fig::memx}
\end{figure}

\emph{MemX} is prototyped using smart eyewear with an inward-facing eye camera and a forward-facing world camera. 
The eye camera performs gaze tracking. The world camera samples and 
analyzes the field of view. Fig.~\ref{fig::memx} illustrates the hardware prototype of \emph{MemX}. 
The first generation of \emph{MemX} uses a Logitech B525 with 1,280$\times$720 resolution as the front-view 
camera installed on the upper side of the eyewear frame, and a Sony IMX 291 with 320$\times$240 resolution 
as the eye camera, secured by a standalone arm. 
We are in the process of developing the second-generation hardware prototype. A key modification is to 
remove the standalone arm to improve the usability and user-friendliness. 

The first-generation prototype uses the Jetson Xavier NX mobile computing platform. We are in the process of 
migrating the \emph{MemX} design to the Ambarella CV28 low-power vision platform directly integrated into one
of the eyewear legs, with a battery in the other leg, similar to the hardware design of Spectacles from Snapchat. 
The CV28 platform is equipped with built-in advanced image processing, high-resolution video encoding, and 
CVflow computer vision processing capabilities~\cite{ambarellacv28}.
The typical power consumption of CV28 is in the range of \unit[500]{mW}, which is suitable for battery-powered wearable design. 
 
\subsection{\emph{MemX} Software Operation}
\emph{MemX} is equipped with the proposed TVA network, which uses attention tracking to \textcolor{black}{achieve personalized moment auto-capture.}
The TVA network continuously tracks eye gaze to detect potential attention events. 
If a possible attention event is detected, the TVA network will turn on the world camera, and eye gaze and 
scene sequences are both fed into the TVA network to make the final attention decision. 

As described in Section~\ref{sctn::method}, attention is a direct indicator of salient visual content. Furthermore,
the attention level, or the level of interest, can be quantified by $T$, the duration of the smooth pursuit phase,
and longer duration implies stronger personal interest. Therefore, $T$ controls 
the selectivity of moment recording. Furthermore, the proposed TVA network uses computation-efficient eye-tracking to detect potential attention events. A larger $T$ makes eye-tracking  more 
selective. Therefore, the fusion stage of the TVA network, which is more computation and energy intensive, will 
be triggered less frequently, thereby improving system energy efficiency. 
The details of how $T$ affects the performance and energy consumption 
of \emph{MemX} are discussed in Section~\ref{sctn::exp}. 
One other design parameter of \emph{MemX} is the duration of each video snippet. In the current implementation, \emph{MemX} continuously records a detected moment till a pre-defined duration threshold is reached. 

\subsection{Energy Model}
The energy consumption of \emph{MemX} is mainly contributed by the world camera, eye camera, the TVA network, and also high-resolution video recording. Next, we characterize the energy consumption of these key components. 

\subsubsection{Energy Consumption of \textcolor{black}{Imaging} Pipelines}
The operation of an imaging pipeline starts with sensing incoming light and converting it into electrical signals. 
Then, an image signal processor (ISP) receives the electrical signals and encodes them into a compressed 
format. The energy consumption of \textcolor{black}{imaging pipeline} is contributed by the following three 
components~\cite{lubanaDigitalFoveationEnergyAware2018}: $E_{\mathit{sensor}}$ (image sensor), $E_{\mathit{ISP}}$ (ISP), and $E_{\mathit{comm}}$ (communication), 
as follows:
\begin{equation}
\label{eq::ec_camera}
E_{\mathit{imaging}} =E_{\mathit{sensor}}+ E_{\mathit{ISP}}+E_{\mathit{comm}}.
\end{equation}

Next, we discuss the energy consumption of the aforementioned components.

\paragraph{(1) Energy consumption of image sensing} The operation of an image sensor consists of three states, i.e., \textcolor{black}{idle, active, and standby}. The power consumption of the standby state is negligible (typically in the range of 0.5-\unit[1.5]{mW}), thus ignored from the energy model. \textcolor{black}{Eq.~\ref{eq::ec_sensor} defines sensor energy, as follows:}
\begin{equation}
\label{eq::ec_sensor}
E_{\mathit{sensor}} = P_{\mathit{sensor, idle}}\cdot T_{\mathit{exp}}+P_{\mathit{sensor, active}}\cdot T_{\mathit{sensor, active}},
\end{equation}
where $P_{\mathit{sensor, idle}}$ and $P_{\mathit{sensor,active}}$ are the average power consumption when the sensor is in the idle state and active state, respectively. $T_{\mathit{exp}}$ is the exposure time, and the image sensor is idle during the exposure phase. $T_{\mathit{active}}$ is the time duration when the image sensor is active, which is determined by the ratio of transferred frame resolution $R_{\mathit{frame}}$ to the external clock frequency $f$, i.e., $R_{\mathit{frame}}/f$. Here,  $P_{\mathit{sensor, idle}}$ and $f$ can be viewed as sensor-specific constants, and $P_{\mathit{sensor,active}}$ is a linear function of sensor resolution $R$ ($R \geq R_{\mathit{frame}}$). 

\paragraph{(2) Energy consumption of ISP} The ISP operates in two states: idle and active. It is active during image processing ($T_{\mathit{ISP}}$) and idle during image sensing. The time for image sensing is the sum of exposure time $T_{\mathit{exp}}$ and the transferring time of frame (in pixels), i.e., $R_{\mathit{frame}}/f$. The energy consumption of ISP is then determined as follows. 
\begin{equation}
\label{eq::ec_isp}
E_{\mathit{ISP}} = P_{\mathit{ISP, active}} \cdot T_{\mathit{ISP}} + P_{\mathit{ISP, idle}} \cdot (T_{\mathit{exp}}+ R_{\mathit{frame}}/f),
\end{equation}
where $P_{\mathit{ISP, active}}$ and $P_{\mathit{ISP, idle}}$ are the average power consumption of the ISP in the active and idle state, respectively.

\paragraph{(3) Energy consumption of communication interface.} The energy consumption of the communication interface $E_{\mathit{comm}}$ is a linear function of the number of transferred frame pixels $R_{\mathit{frame}}$~\cite{lubanaDigitalFoveationEnergyAware2018}, as follows:
\begin{equation}
\label{eq::ec_comm}
E_{\mathit{comm}} = k \cdot R_{\mathit{frame}},
\end{equation}
where $k$ is a design-specific constant determined by the communication interface.

As can be seen from Eq.~\ref{eq::ec_camera}, \ref{eq::ec_sensor}, \ref{eq::ec_isp}, and \ref{eq::ec_comm}, 
the energy consumption of cameras highly depends on the sensor resolution. For \emph{MemX}, the energy 
consumption of the high-resolution world camera is significantly higher than that of the eye camera. 
Therefore, the TVA network uses eye-tracking alone to first detect potential attention events, and only
then turns on the world camera for accurate attention detection and salient content recording, thereby 
effectively minimizing the energy cost \textcolor{black}{from} the world camera. 

\subsubsection {Energy Consumption of \emph{MemX}}
\textcolor{black}{\emph{MemX} is equipped with the proposed TVA network.}
The operation of the TVA network consists of two stages. First, the TVA network continuously performs 
eye-tracking to detect potential attention events. This stage has high energy efficiency, thanks to 
the low data rate of the eye camera and \textcolor{black}{computationally-efficient eye tracking}. When a possible 
attention event is detected, the TVA network invokes the second stage, using a light-weight network 
to perform eye-scene feature fusion \textcolor{black}{in order to finalize the attention decision.}
We denote the average power of the two stages as $P_{\mathit{eye \;tracking}}$ and $P_{\mathit{fusion}}$, respectively.  

\textcolor{black}{The energy consumption of \emph{MemX} is then formulated as follows:}
\begin{align}
E_{\mathit{MemX}}&= T_{\mathit{always-on}} \times (P_{\mathit{eye \;camera}} + P_{\mathit{eye\; tracking}})  + T_{\mathit{fusion}} \times (P_{\mathit{world \;camera}} + P_{\mathit{fusion}} )  \nonumber \\
&+ T_{\mathit{auto-captured}} \times (P_{\mathit{world \;camera}}  +P_{\mathit{encoding-storing}} ),
\label{eq::ec_host}
\end{align} 
\textcolor{black}{where $T_{\mathit{always-on}}$ is the operation time of \emph{MemX}, $P_{\mathit{eye \; camera}}$ and $P_{\mathit{world \; camera}}$ are the power consumption of the inward-facing eye camera and the forward-facing world camera, respectively, $T_{\mathit{fusion}}$ is the operation time of eye-scene feature fusion, $T_{\mathit{auto-captured}}$ is the operation time of high-resolution video recording when human visual attention events are detected,  $P_{\mathit{encoding-storing}}$ is the power consumption of the host processor during high-resolution video recording, mostly contributed by video encoding and storage.}
Compared with the eye-tracking alone stage, the second stage is more data and computation intensive,  
but is only triggered when potential attention events are detected,
and stays off most of the time, thus effectively reducing energy cost. In addition, the TVA network
consists of a new light-weight network architecture, which is significantly more efficient than the
existing VIS-based design, e.g., MaskTrack R-CNN architecture~\cite{yang2019video}. As shown in the experimental
studies, the proposed TVA network significantly outperforms the VIS-based method in terms of system
energy efficiency. 

\section{Evaluations}
\label{sctn::exp}
This section evaluates \emph{MemX}, the proposed attention-aware smart eyewear system for personalized moment auto-capture. 
We first detail the experimental settings, including user studies, data collections, and evaluation metrics. 
Then we present the evaluation results to quantify the performance and efficiency of \emph{MemX}. The studies 
described in this section are conducted in a controlled lab environment. In-field pilot studies will be presented 
in the next section.

\subsection{Evaluation Methodology}
\subsubsection{Experimental Setup}
\begin{figure*}[ht]
    \includegraphics[width=1.0 \textwidth]{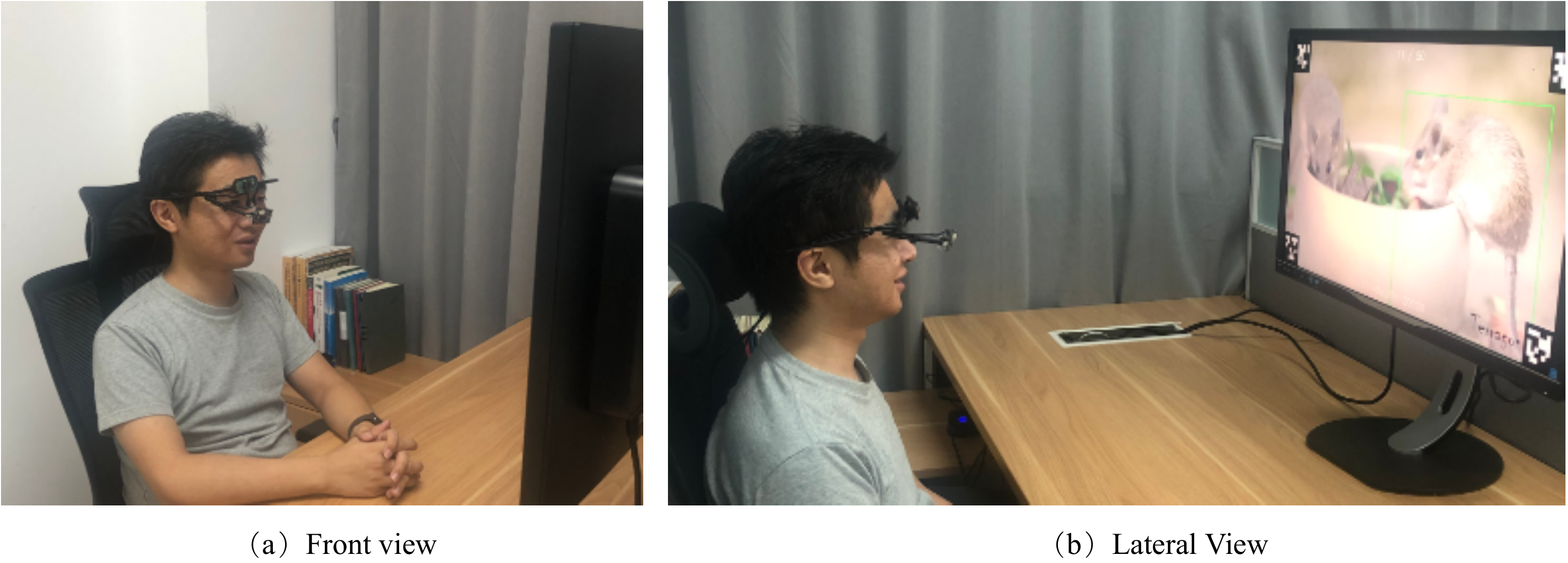}
    \caption{\emph{MemX} experimental study: The controlled lab setting.}
    \label{fig::wearing_memx}
\end{figure*}

The controlled lab setting is shown in Fig.~\ref{fig::wearing_memx}. A group of participants wear \emph{MemX}, 
sit in front of a computer, and watching a sequence of video clips. During each study, 
video clips are recorded through the field of view of \emph{MemX}, which simultaneously tracks the participant's visual attention. In total, 30 participants are recruited. A summary of the participants are provided below.
\begin{itemize}
 \item \textbf{Gender}: 10 (33.3\%) female and 20 (66.7\%) male,
 \item \textbf{Age}: between 21 and 35,
 \item \textbf{Vision}: normal vision (7, 23.3\%), nearsighted $\leq$ 3.0 diopters (8, 26.7\%) and $\ge$ 3.0 diopters (15, 50.0\%).
\end{itemize}

\subsubsection{Video Data Preparation}
In this work, we adopt the YouTube-VIS dataset\footnote{\url{https://youtube-vos.org/dataset/vis/}}~\cite{yang2019video} as ``micro-benchmark in a controlled setting'' to quantitatively evaluate such technical capability of \emph{MemX}. In addition, we use pilot studies to further evaluate \emph{MemX} in real-life scenarios.

Specifically, the YouTube-VIS dataset covers a wide range of complex real-world scenarios. For instance, the dataset consists of temporally dynamic scenes, such as sports, as well as spatially diverse scenes, such as multiple objects within the same scene. Therefore, using the YouTube-VIS dataset, we can design a wide range of interesting testing cases to evaluate the accuracy and efficiency of \emph{MemX} for eye tracking and attention detection. For example, the participant is initially attracted by a white duck within a scene consisting of multiple animals, and then his or her attention shifts towards a different animal. In addition, as a widely used dataset, the YouTube-VIS dataset includes necessary object annotation information regarding object location, classification, and instance segmentation. Such annotations help us to accurately set up the ground truth in terms of the objects of interest during the experiments.

\begin{figure*}[ht]
    \includegraphics[width=1 \textwidth]{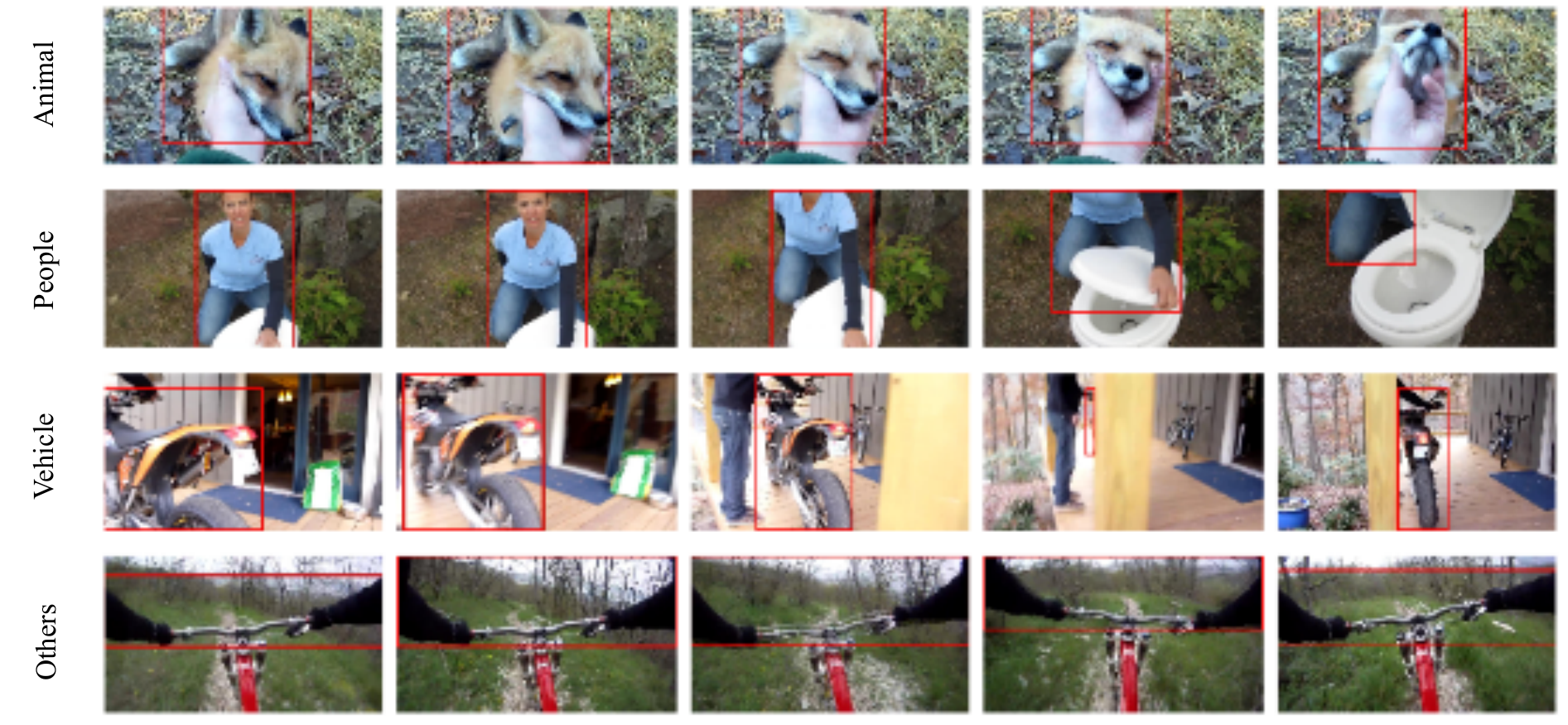}
    \caption{YouTube-VIS dataset illustration. The bounding boxes mark the pre-selected objects of interest.}
    \label{fig::youtube_vis}
\end{figure*}

The YouTube-VIS dataset consists of a 40-category label set and 2,238 videos with released annotations. Each video snippet 
lasts 3 to 5 seconds with a frame rate of \unit[30]{fps}, and every 5th frame is annotated for each video snippet. 
We first divide the 2,238 videos into four classes based on their content, including \emph{Animal}, \emph{People}, 
\emph{Vehicle}, and \emph{Others}. As a result, the four classes  contain 1487, 437, 215, and 99 videos, respectively. 
We then randomly select videos in each class, resulting in 1,000 videos total (---\emph{670}, \emph{197}, \emph{97}, and \emph{36}), which are used in 
this study. Fig.~\ref{fig::youtube_vis} visualizes cases of the 4-class video frames in YouTube-VIS dataset.

Since each video snippet lasts 3 to 5 seconds in the YouTube-VIS dataset, we concatenate multiple video snippets to form videos with a duration of about 7-15 minutes (approximately 100 video snippets). The concatenated video is then provided to the participants to watch. We recruited 30 volunteers to collect their eye data when they watch the concatenated videos.
To alleviate the potentially person-dependent bias, we ensure each video snippet is watched by several participants (three in this work). Given the 30-recruited participants, we select 1,000 video snippets from the entire YouTube-VIS datasets.

\subsubsection{Eye Data Collection}
\emph{MemX} uses the inward-facing eye camera to capture eye video data. We calibrate \emph{MemX} to correlate \Note{*** RD: Do you mean ``correlate'' or ``synchronize''? *done by YZ*}
the world camera with the eye camera. The calibration method follows Pupil 
Capture\footnote{\url{https://docs.pupil-labs.com/core/software/pupil-capture/\#calibration}}~\cite{kassner2014pupil}. Each participant watches approximately 100 videos randomly selected from the four classes in our benchmark video data. That ensures each video in our benchmark can be watched by three participants. We pre-select the target attentive object in each video snippet as users' visual interest, and we guide participants to gaze at the pre-assigned object and track their motion. We then obtain the users' eye video dataset that is \textcolor{black}{synchronized} with the video dataset.
\Note{***RD: Synchronized? *done by YZ: YES*}

In the following experiments, the video dataset and the eye dataset are randomly divided into training set, test set, 
and validation set with a 70\%:10\%:20\% ratio.

\subsubsection{Evaluation Metrics}
We use precision and recall to evaluate the accuracy of human visual attention tracking of \emph{MemX}, defined as follows:
\begin{align}
\label{eq::precision}
\mathit{precision}  &= \frac{\mathit{TP}}{\mathit{TP}+\mathit{FP}} \text{ and } \\
\label{eq::recall}
\mathit{recall}  &= \frac{\mathit{TP}}{\mathit{TP}+\mathit{FN}},
\end{align}
where $\mathit{TP}$ (true positive) denotes when \emph{MemX} correctly identifies the attentive object of interest, 
$\mathit{FN}$ (false negative) denotes when \emph{MemX} fails to identify the attentive object of interest, 
and $\mathit{FP}$ (false positive) denotes when \emph{MemX} incorrectly identifies an object of interest that 
the user actually did not pay attention to. In summary, the higher the precision and recall, 
the better the accuracy of \emph{MemX}. 
In addition we also consider average precision (AP) which jointly considers precision and recall measures, as follows. 
\begin{equation}
\label{eq::map}
\mathit{AP}  = \sum_n (\mathit{recall}_n- \mathit{recall}_{n-1})\times \mathit{precision}_n,
\end{equation}
where $\mathit{precision}_n$ and $\mathit{recall}_n$ are the precision and recall at the $n$th threshold. 

To evaluate the energy efficiency of \emph{MemX}, we consider the following hardware settings. 
The \emph{MemX} prototype is equipped with a Sony IMX 291 based eye camera and a Logitech B525 based world camera. 
The Logitech B525 camera supports a maximum resolution of \unit[2.07]{M-pixels}.
In addition, we target the Ambarella CV28 low-power computer vision SoC platform, which is equipped with 
advanced image processing, high-resolution video encoding, and CVflow computer vision processing 
capabilities. The power estimation is based on the model described in Section~\ref{sctn::sys}. 

\subsubsection{Training TVA Network} 
This work uses Adam~\cite{kingma2014adam} as the optimizer, with an initial learning rate of 0.01, which is decreased by 10\%
after every 30 epochs until 100 epochs have occurred. For the shallow network $\mathcal{N}_{\mathit{SLM}}$ in TVA, we adopt the first two blocks of MobileNetV2 to extract the scene frame features as MobileNetV2 is efficient enough for mobile devices~\cite{sandler2018mobilenetv2}. We use pre-trained weights provided in prior work~\cite{sandler2018mobilenetv2} as a starting point for training TVA. During training, we freeze $\mathcal{N}_{\mathit{SLM}}$ during the first 30 epochs.


\subsubsection{Baselines}
To our best knowledge, there is no prior work that targets the same research challenge and is directly comparable to this work. 
For comparison, we consider the following two baselines. 

\paragraph{(1) \textcolor{black}{Eye-tracking-alone.}} This approach uses eye tracking alone to capture potential attentive visual content. 
Specifically, we first use eye tracking to detect \textcolor{black}{the} saccade-smooth pursuit transition, an indication of potential visual attention shift. Then, we measure the gaze regional focus relative to the field of viewing during a smooth pursuit phase or a fixation phase for a time period.
We experimentally define the occurrence of actual attention as when 90\% gazes are located in a close region with area 
$A = (0.05\times W) \times (0.05\times H)$ for a time period $t$, where $W$ and $H$ denote the width and height of viewing scene frame, 
respectively, and 0.05 is the rescaling ratio. Considering that the Logitech B525 camera has a 69$^{\circ}$ diagonal 
field-of-view~\footnote{\url{www.logitech.com/assets/64667/b525-datasheet.ENG.pdf}} and angular error is approximately with 
median value of 3.45$^{\circ}$~\cite{cazzato2020look}, we estimate the rescaling ratio as 3.45/69 (i.e., 0.05)  
in this work. The effectiveness of how $T$ affects the performance of attention detection is evaluated in Section~\ref{sctn::usability}.
The following experiments use the eye-tracking-alone method to establish the baseline accuracy of the proposed work. 

\paragraph{(2) VIS-based method.} This method uses eye tracking and VIS-based object detection to jointly capture potential 
attentive visual content, and the VIS-based object detection task adopts the MaskTrack R-CNN architecture~\cite{yang2019video}. 
Different from the proposed TVA network, in the VIS-based method, eye tracking and VIS-based object detection are two independent, parallel tasks. As a result, VIS-based object detection is always on, which may introduce significant 
computation and energy overhead. As will be shown in the experimental results, the processing speed of the VIS-based method
is only 0.6 frames per second, preventing it from practical adoption into wearables. The purpose of including this 
method in the experiments is to establish the baseline energy efficiency of the proposed work.

\paragraph{(3) Saliency-map-based method.} 
The saliency detection method \cite{Marat2009Modelling} aims to predict the salient regions that has potentially attracted the use's attention, which is somehow similar to the target of the proposed TVA network. For a fair and complete comparison, we adopt and evaluate the saliency detection method in \cite{Marat2009Modelling}
as an additional baseline. We will refer it as the saliency-map-based method for simplicity.

\subsection{Results}

\subsubsection{Overall Performance}
Table~\ref{tb::overrall_performance} shows the accuracy and energy efficiency comparison between \emph{MemX} and the
three baseline methods. The precision, recall, and average precision AP of each method are shown in columns 2, 3 and 4. 
The energy reduction of the proposed work and the eye-tracking-alone 
based baseline method relative to that of the VIS-based baseline method is shown in column 5. 

As the table shows, the eye-tracking-alone method achieves low precision and recall: it cannot
be used for accurate and stable attention tracking. In contrast, the proposed TVA-based method significantly improves the attention tracking precision and recall, with only a slight increase in energy consumption. 
In addition, the proposed TVA network achieves higher average precision than the two baseline methods. 

Among these four methods, the VIS-based method achieves the highest precision and recall, which however is associated with 
significant energy and computation penalties. Specifically, compared with the proposed TVA-based method, the energy
consumption of the VIS-based method is 2.7$\times$ higher. Furthermore, using the Jetson Xavier NX platform, the processing
speed of the VIS-based method is 0.6 frames per second, which is too slow for practical use. In contrast, the
processing speed of the proposed TVA-based method exceeds 30 frames per second, offering at least a 50$\times$ speedup. 

\textcolor{black}{Compared with the VIS-based method, the saliency-map-based method is with significant accuracy degradation. This is due to the fact that the saliency-map-based method can only provide region-level information, instead of object-level information. An object often contains multiple regions, and human's visual attention on a specific object may temporally shift across the multiple regions of the same object. As a result, the saliency-map-based method fails to accurately detect whether a person gazes at the same object. Compared with the saliency-map-based method, the VIS-based method offers object-level information, thus offering better visual attention analysis accuracy. Also, the proposed TVA-based method outperforms the saliency-map-based method. This is because the TVA network is designed to provide object-level information in a light-weight architecture complemented with the regional-level eye-tracking task. Experimental results demonstrate that, compared against the saliency-map-based method, the proposed TVA-based method achieves 58.48\% accuracy improvement (average precision) with 45.64\% energy savings gain.}

In summary, the three baseline methods suffer from either serious accuracy or energy efficiency limitations. In 
comparison, the proposed TVA network strikes a good balance between accuracy and efficiency for attention tracking.

  \begin{table}[!htb]
\caption{Accuracy and energy efficiency comparison.} 
\label{tb::overrall_performance}
\begin{tabular}{@{}cccccll@{}}
\toprule
Method &Precision & Recall & Average Precision &  Energy Savings  \\ \midrule
TVA-Based Method (Proposed)     &   87.50\%       &86.40\%  & 93.65\%  &  70.61\% \\
\textcolor{black}{Eye-Tracking-Alone} Method    &    40.52\%       &  12.70\% & 35.93\%   &  76.97\% \\
VIS-Based Method    &     98.08\%    & 90.26\%  & 93.40\% &  0.00\%\\ 
\textcolor{black}{Saliency-Map-Based Method}   &  40.08\%    & 37.94\%  & 35.17\% &  24.97\%\\ \bottomrule
\end{tabular}
\end{table}

\subsubsection{Usability Discussion}
\label{sctn::usability}
\emph{MemX} uses attention tracking to control personalized moment auto-capture. As described 
in Section~\ref{sctn::sys}, the duration of the smooth pursuit phase indicates the level of interest. 
In \emph{MemX}, the proposed TVA network uses the parameter $T$ to determine how long a potential 
smooth pursuit phase has been detected using eye tracking before triggering high-precision attention 
tracking and salient content recording. By adjusting the value of $T$, \emph{MemX} controls the 
selectivity of moment recording, as well as the energy efficiency of the system. 

\begin{figure*}[ht]
\begin{minipage}[t]{0.45\linewidth}
\subfloat[$T$ vs. trigger rate $\alpha$]{\includegraphics[width=1 \textwidth]{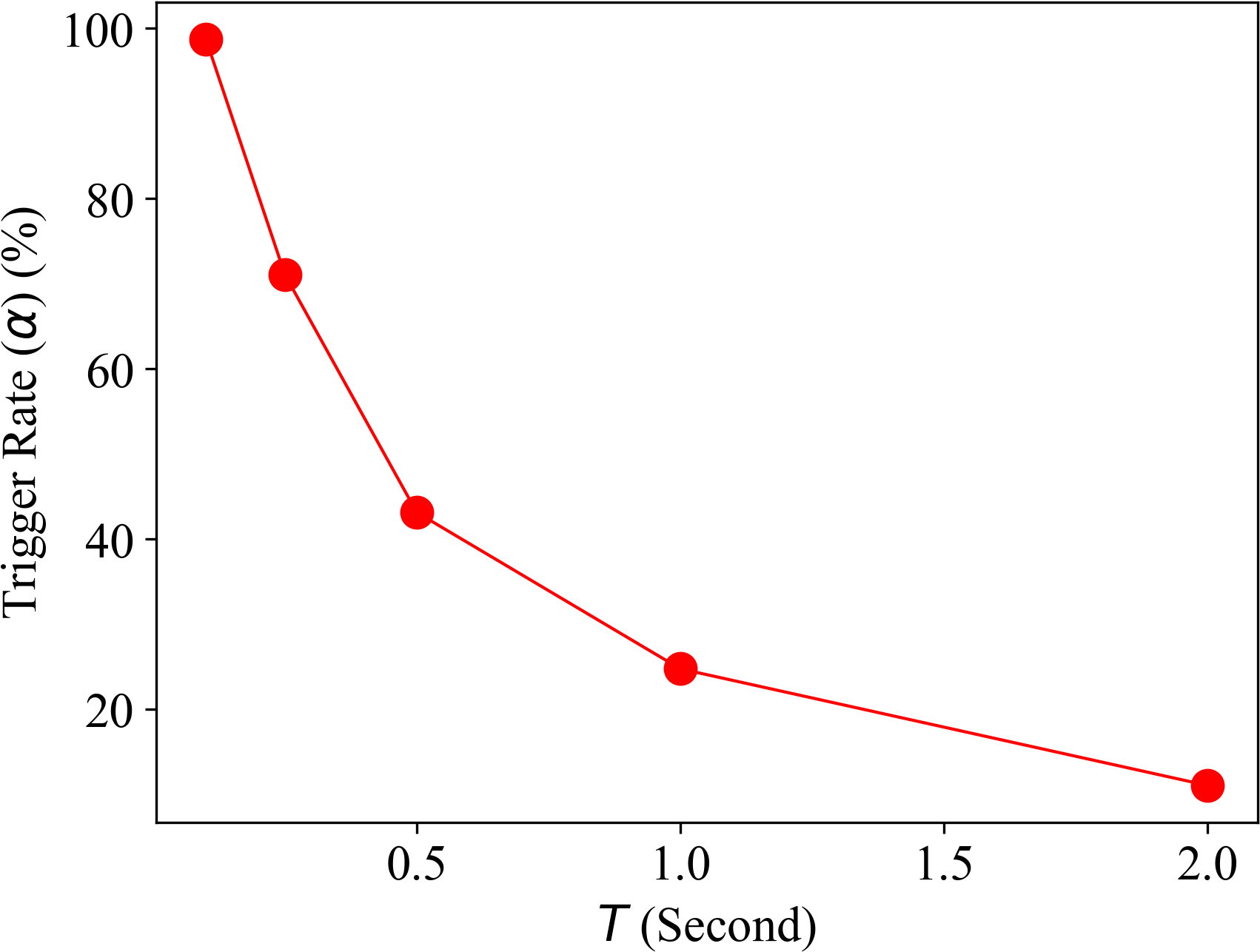}
\label{fig::t_trigger_rate}}
\end{minipage}
\hfill
\begin{minipage}[t]{0.45\linewidth}
\subfloat[$T$ vs. energy savings]{\includegraphics[width=1 \textwidth]{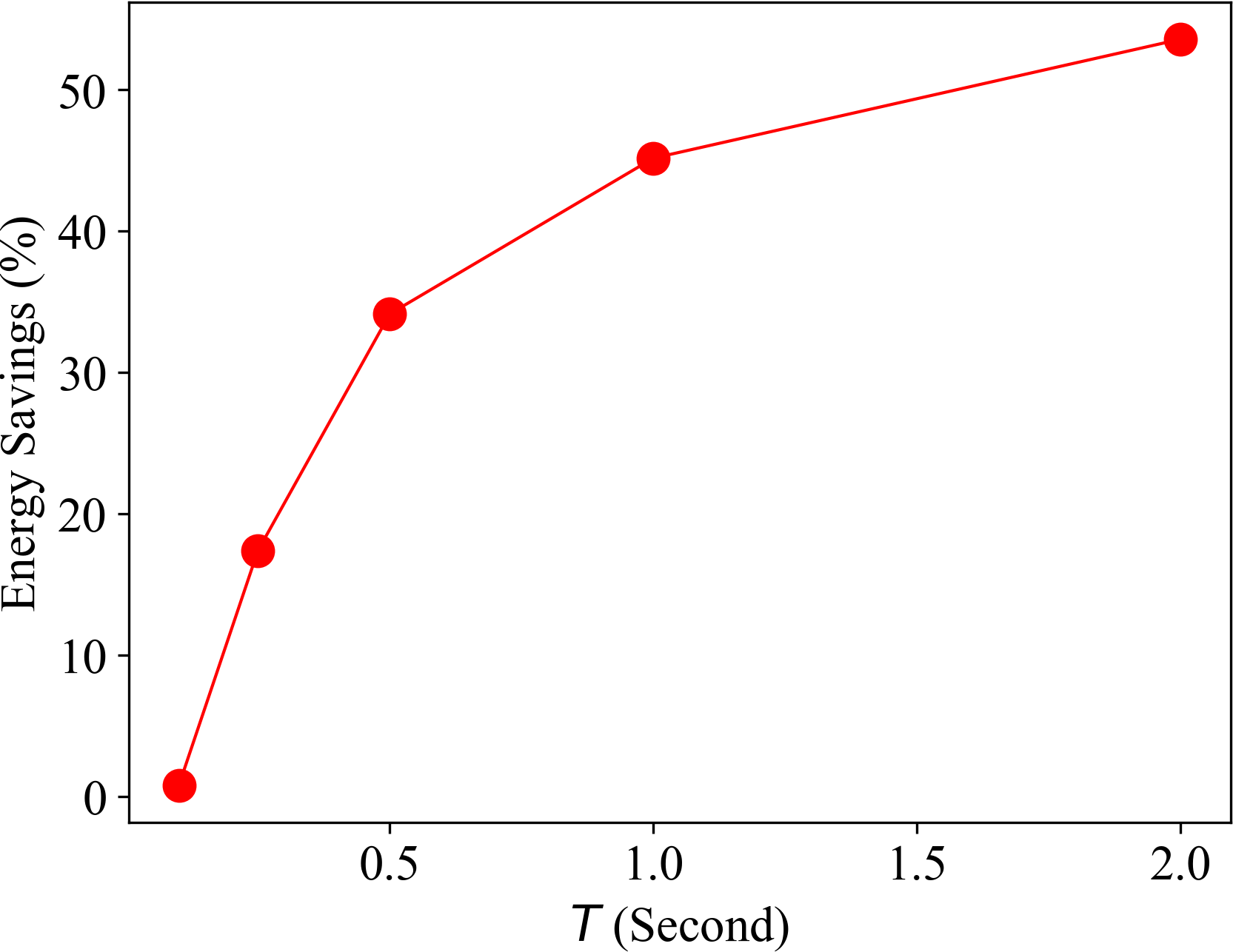}
\label{fig::t_ec}}
\end{minipage}%
\caption{Effects of $T$ on trigger rate $\alpha$ and energy savings. } 
\end{figure*}

Fig.~\ref{fig::t_trigger_rate} shows the impact of $T$ on the trigger rate $\alpha$ of \emph{MemX}. 
As shown in this figure, the trigger rate $\alpha$ decreases as $T$ increases. 
Fig.~\ref{fig::t_ec} shows the impact of $T$ on the energy consumption of attention detection. 
As shown in this figure, the energy saving increases as $T$ increases. This is due to the fact
that, with the increase of $T$, the eye-tracking component of the TVA network becomes more selective, 
and the data and computation intensive fusion stage is less likely to be triggered, thereby reducing 
the energy cost. 

For the in-field pilot studies (described in the next section), $T$ is empirically set to \unit[1]{s}. Based on the 
feedback from the pilot studies, this $T$ value strikes a good balance between selectivity of salient content
recording and system energy efficiency. 

\subsubsection{Case Study}
Next, we present two cases to offer further insights regarding why the proposed TVA-based method outperforms the eye-tracking-alone method for attention detection. 

\begin{figure}[!htb]
	\includegraphics[width=1.0 \textwidth]{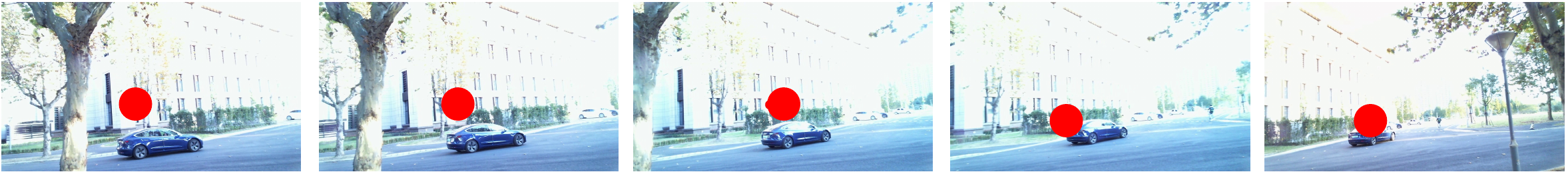}
	\caption{A false negative case illustration.} 
	\label{fig::case_study_fn}
\end{figure}

\paragraph{(1) A False Negative Case.}
As described in Section~\ref{sctn::intro}, eye-tracking-alone cannot provide accurate attention tracking, due to limited tracking 
resolution and inherently noisy eye movement patterns. Fig.~\ref{fig::case_study_fn} shows such a case. As we can see, the user's attention
focuses on a fast-moving vehicle, and gaze trajectory is jittering and noisy. Therefore, it is challenging for eye tracking to 
distinguish between saccade and smooth pursuit. In contrast, the TVA network leverages both regional information provided
by eye tracking and object and semantic level information by video analytics, thus is able to accurately detect the fast-moving vehicle as 
the object of interest. \Note{*** RD: Couldn't this problem be fixed with a simple filter, e.g., a Kalman filter? Why is your approach needed?}

\begin{figure}[!htb]
	\includegraphics[width=1.0 \textwidth]{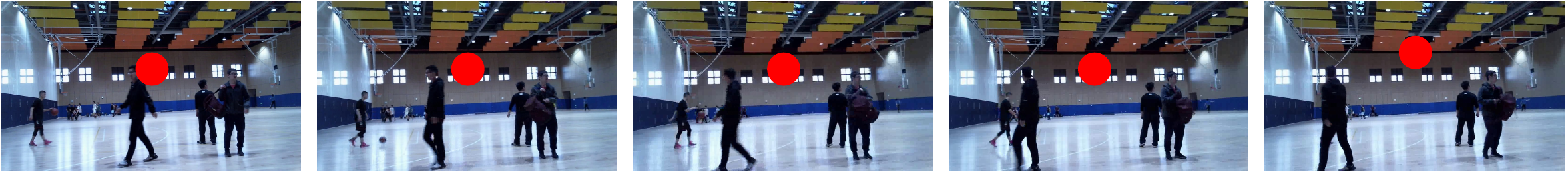}
	\caption{A false positive case illustration.}
	\label{fig::case_study_fp}
\end{figure}

\paragraph{(2) A False Positive Case.}
Consider a the frequent inattentive event in which a person is not paying attention to the surrounding environment and \textcolor{black}{gaze} is fixed on one position. Eye tracking alone will mistakenly classify this case as a qualified attention event. In contrast, 
the TVA network is able to detect such a false positive case. Specifically, as shown in Fig.~\ref{fig::case_study_fp}, the TVA network recognizes that the regional visual focus along the gaze trajectory does not match a specific object. Therefore, 
there is no salient visual object that draws the user's attention. 

\section{Pilot Study}
\label{sctn::app}
\emph{MemX} can be applied to a wide range of potential application scenarios. We have conducted multi-round 
user interviews to explore potential usage scenarios. Suggested popular scenarios include
daily lifelogging, traveling and sightseeing, sports logging, public event summary, etc. After examining these 
scenarios, we observe that different scenarios exhibit distinct characteristics. In particular, we have identified 
two key attributes, namely scene complexity and human attention dynamics to categorize individual scenarios as follows.
\begin{enumerate}

\item \textbf{Scene complexity} depends on the number of objects in each scene as well as their
motion patterns. A simple scene might contain few, mostly stationary, objects while complex scene might contain many fast-moving objects.

\item \textbf{Human attention dynamics} depends on the frequency that human attention switches among objects. For example, in a basketball game, human attention is highly dynamic. Book reading, on the other hand, is mostly stationary.

\end{enumerate}

\begin{figure}[]
	\includegraphics[width=1.0 \textwidth]{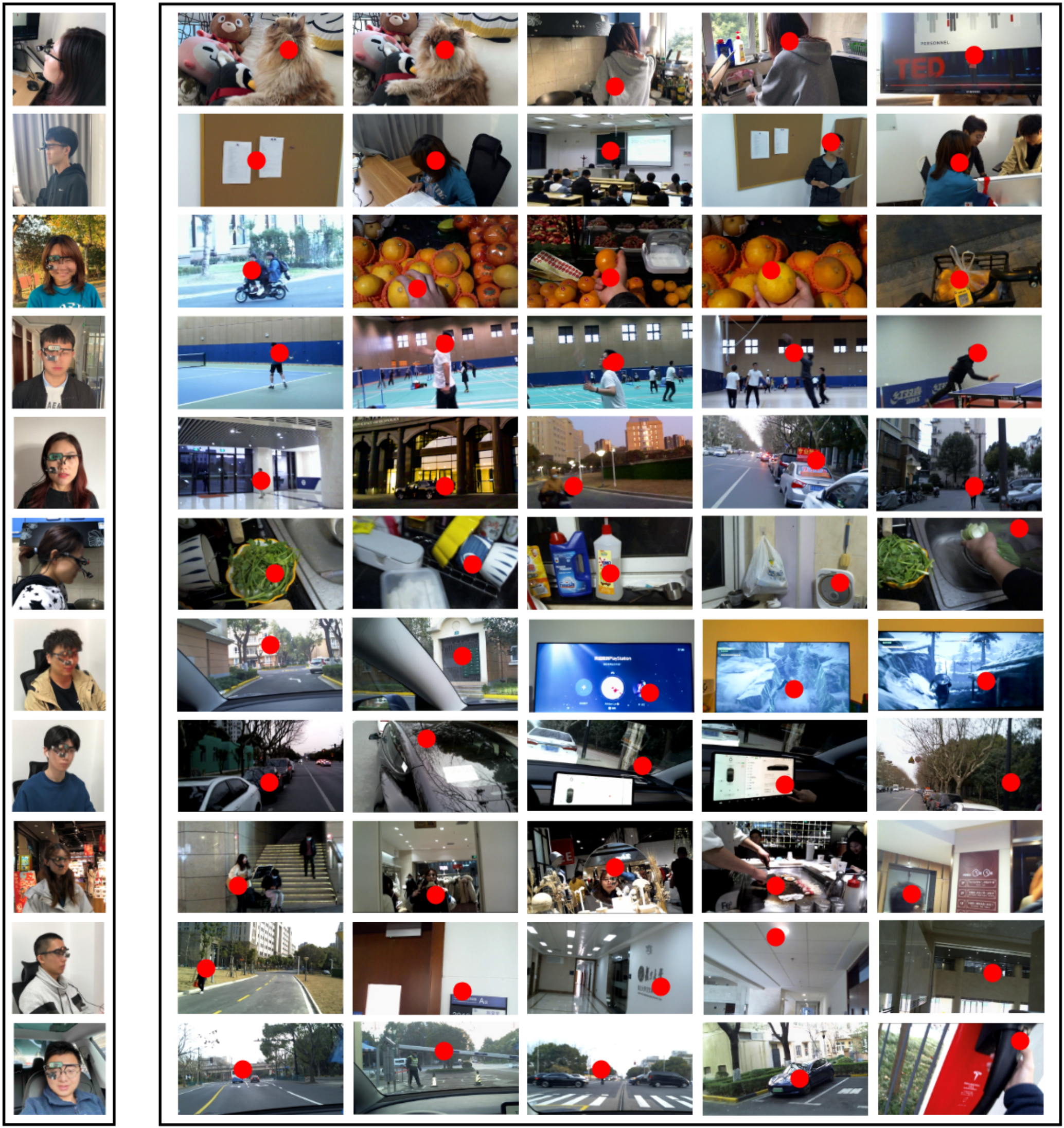}
 \caption{\textcolor{black}{Exemplary moment snapshots captured by \emph{MemX} in the 11 pilot studies. The first column shows the 11 volunteers wearing~\emph{MemX}. The rest are the captured moment snapshots. The red circle indicates the measured visual attention by~\emph{MemX}. The 11 studies, from top to bottom, are (1) lifelogging, (2) academic events, (3) grocery shopping, (4) sports logging, (5) going for a walk, (6) cooking, (7) playing game, (8) driving \#1, (9) shopping mall, (10) on campus, and (11) driving \#2.}}
 	\label{fig::applications}
 \end{figure}

\subsection{Data Collection and Description}
\emph{MemX} aims to enable automated, personalized capture of interesting visual content, with energy-constrained wearable form factor. To evaluate the accuracy regarding personalized moment auto-capture and energy efficiency of \emph{MemX}, we select a set of potential usage scenarios, covering the four combinations of the two characteristics. We then conduct in-field pilot studies targeting these scenarios, including 11 pilot studies in total (5 female, 6 male, aged between 21 and 40), covering a diverse range of real-life scenarios. During each study, the participants wear \emph{MemX} to auto-capture visual moments of interest in the form of short video clips. For comparison purposes, the complete visual experience of each study is also recorded by the world camera of \emph{MemX} (baseline video). In total, equipped with \emph{MemX}, 11 participants recorded approximately 245 minutes of baseline video, with about 37 minutes of video clips auto-captured as moments of interest. Fig.~\ref{fig::applications} shows some exemplary moment snapshots captured by \emph{MemX}.

\subsection{Accuracy of \emph{MemX} in Pilot Study}

\subsubsection{Overall Accuracy}
To evaluate the accuracy of \emph{MemX}, after each user study, we asked the participant to review the baseline video and manually mark the moments reflecting his or her true interest during recording (ground truth), which are then compared against the video clips auto-captured by \emph{MemX}. Table~\ref{tb::pilot_study} summarizes the accuracy of \emph{MemX}. As shown in Table~\ref{tb::pilot_study}, \emph{MemX} can accurately detect and automatically capture 96.05\% of visual moments of interest across the 11 pilot studies. In other words, the video clips auto-captured by \emph{MemX} accurately reflect the participants' moments of interest.

\begin{table}[h]
\caption{Summary of the in-field pilot study.}
\label{tb::pilot_study}
\begin{tabular}{@{}ccccccccll@{}}
\toprule
No. &Scenario & \tabincell{c}{Record\\ (minute)}& \tabincell{c}{Auto-Captured \\ (minute)}&\tabincell{c}{Duration\\ Reduction}& \tabincell{c}{Number of \\True Interest} & \tabincell{c}{Number of \\Missed Interest} & \tabincell{c}{Energy\\Savings}\\ \midrule
1  &  Lifelogging     &  1.76 & 0.13   & 92.61\% & 3 & 0 & 90.68\%\\
2  &  Academic events     &  4.72 & 0.67 & 85.74\% & 5 & 0 & 83.80\%\\
3  &  Grocery shopping     &  3.71 & 0.47 & 87.41\% & 3 & 0 & 85.48\%\\
4  &  Sports logging     &  5.12 & 0.29 & 94.37\% & 3 & 0 & 92.43\%\\
5  &  Going for a walk  & 9.01 & 1.14 & 87.30\% & 5 & 0 & 85.37 \% \\
6  &  Cooking      &  30.78 & 6.28 & 79.61\%  & 6 &  0 & 77.67 \% \\
7  &  Playing game   &  22.74 & 0.44 & 98.08\% & 5 & 2 & 96.15 \% \\
8  &  Driving \#1  &  32.49 & 4.69 & 85.58\% & 5 & 1 & 83.64\% \\
9  &  Shopping mall &  71.24 & 18.94 & 73.42\% & 22 & 0 & 71.49\% \\
10 &  On campus & 31.82 & 0.96 & 96.99\% & 10 & 0 & 95.06\% \\
11 &  Driving \#2 & 31.15 & 3.07 & 90.15\%  & 6 & 0 & 88.22\% \\
\bottomrule
\end{tabular}
\end{table}

\emph{MemX} analyzes the relationship between human attention and interest from the following three aspects simultaneously: (1) the temporal transition from saccade to smooth pursuit, which suggests potential visual attention shift; (2) the gaze duration of following a moving target or fixating on a stationary target, which qualitatively measures the current interest level. In general, the longer the duration, the more interested the user might be; and (3) scene analysis and understanding, which helps to detect cognitively whether there are potentially interesting objects within the region of gaze points. 
By jointly considering the aforementioned three aspects, \emph{MemX} is able to tackle special corner cases such as mind-wandering and driving scenarios. For example, the user may gaze at a position for a long time unconsciously when mind wandering. In this case, \emph{MemX} leverages scene understanding to help decide whether a potential target of interest exists or not. If not, \emph{MemX} filters out those moments. In another case, the user's attention may temporally drift away and then quickly shift back if no interesting objects are detected. Such attention shifts can be captured and then discarded (due to short duration) by \emph{MemX}. We have included pilot studies for these corner cases. The pilot studies demonstrate that \emph{MemX} can successfully filter out these corner cases and capture the true moments of \textcolor{black}{interest}.

\subsubsection{Two Exemplary Cases}
\begin{figure}[]
	\centering
	\begin{minipage}[t]{1.0\linewidth}
		\includegraphics[width=1 \textwidth]{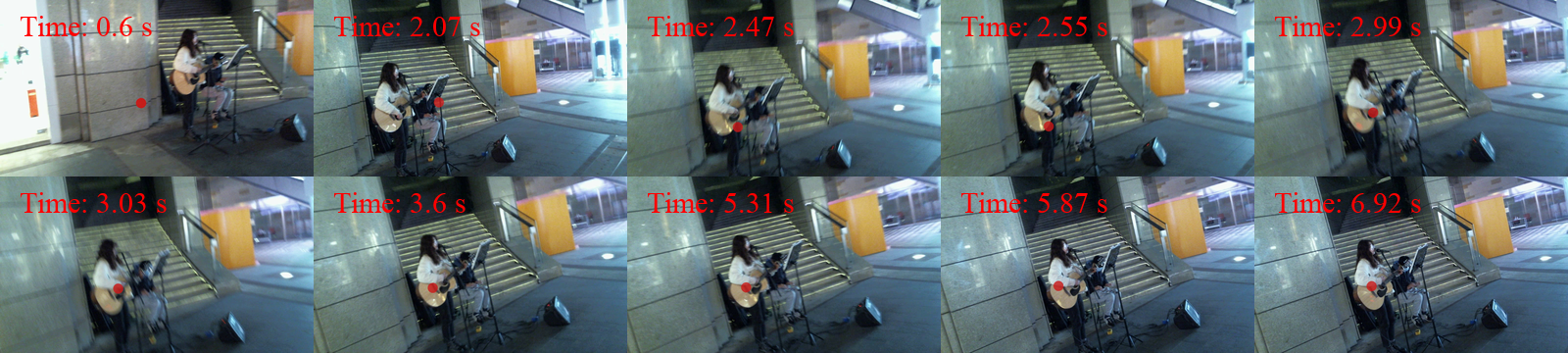}
\caption{Examples of video frames with marked gaze points (red circle) when a participant is enjoying a busker show scenario.}
\label{fig::tp_world}
	\end{minipage}
	
	\begin{minipage}[t]{1\linewidth}
	\includegraphics[width=1\textwidth]{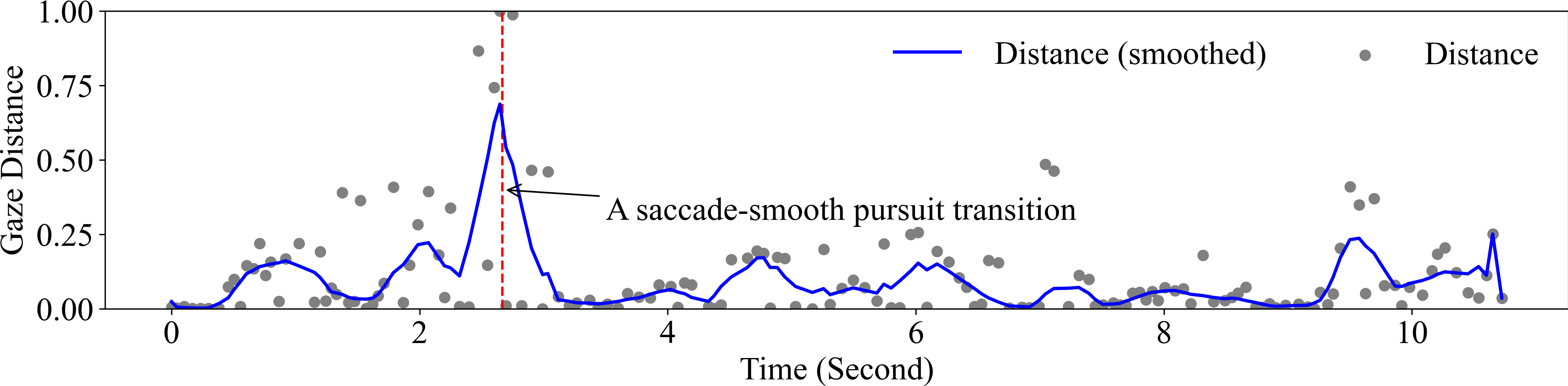}
    \caption{An example of time-series gaze distance variation when a participant is enjoying a busker show scenario.}
    \label{fig::tp_gaze}
	\end{minipage}	

	\begin{minipage}[t]{1.0\linewidth}		\includegraphics[width=1 \textwidth]{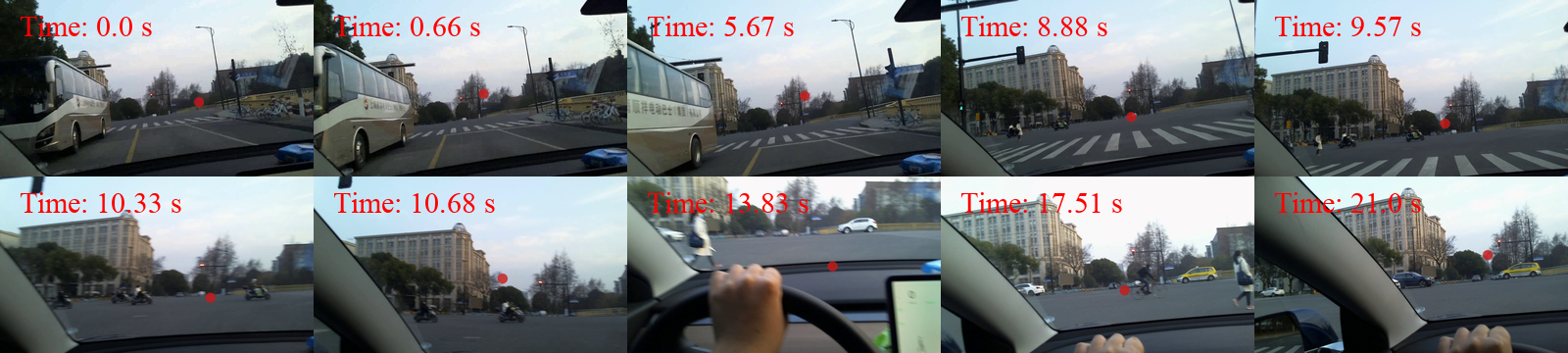}
\caption{Examples of video frames with marked gaze points (red circle) when a participant is driving.}
\label{fig::tn_drive}
	\end{minipage}
	
	\begin{minipage}[t]{1\linewidth}
	\includegraphics[width=1\textwidth]{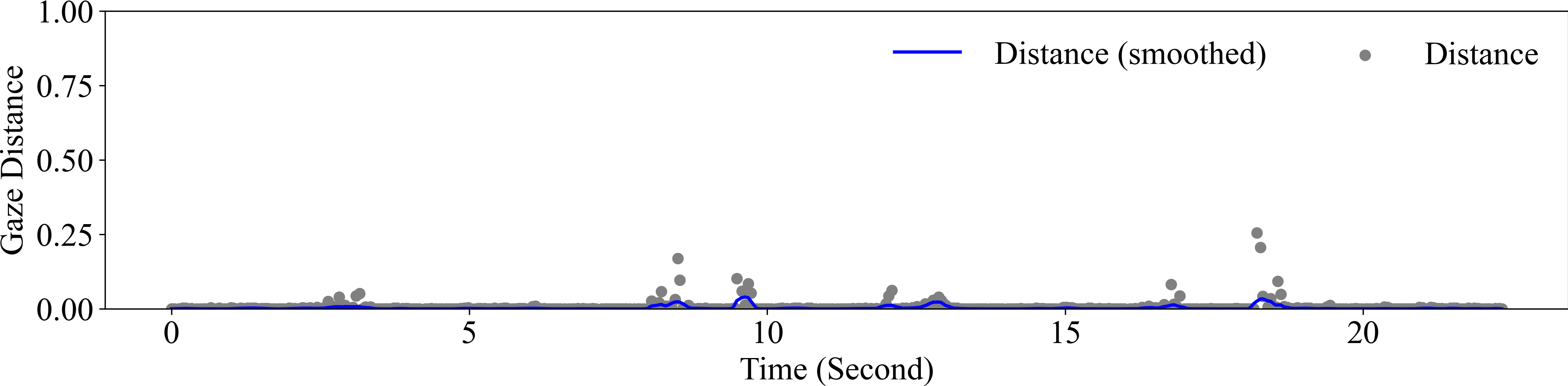}
    \caption{An example of time-series gaze distance variation when a participant is driving.}
    \label{fig::tn_gaze}
	\end{minipage}	
\end{figure}

The following two exemplary cases provide further intuition. Fig.~\ref{fig::tp_world} and Fig.~\ref{fig::tn_drive} show true moments of interest auto-captured by \emph{MemX} and otherwise discarded by \emph{MemX}, respectively.  
Specifically, Fig.~\ref{fig::tp_world} and Fig.~\ref{fig::tn_drive} show successive video frame sequence recorded by \emph{MemX} with marked gaze positions (red circle). Fig.~\ref{fig::tp_gaze} and Fig.~\ref{fig::tn_gaze} illustrate the time-series normalized gaze distance between two successive frames. For the true moments of interest shown in Fig.~\ref{fig::tp_world} and Fig.~\ref{fig::tp_gaze}, we can observe that: (1) there is a saccade-smooth pursuit transition at approximately \unit[2.40]{seconds}; (2) after that,  most of the eye movements are smooth pursuit or fixation; and (3) the attention is located on a girl who is playing guitar. In a driving case, as shown in Fig.~\ref{fig::tn_drive}, the driver gazes at a fixed location in his field-of-view without any obvious target or object. As shown in Fig.~\ref{fig::tn_gaze}, we can see some gaze shifts, such as at time \unit[17]{seconds}. However, those gaze shifts do not qualify for possible attention because we cannot find one stable object that the user continuously focuses on. Thus, those moments are discarded by \emph{MemX}.

\subsection{Energy Efficiency of \emph{MemX} in Pilot Study}
Energy efficiency is essential to wearable devices. In \emph{MemX}, high-resolution video capture and content recording pipeline through the world camera is energy demanding. \emph{MemX} significantly reduces such use only when potential visual attention and moments of interest are detected. As shown in Table~\ref{tb::pilot_study}, the pilot studies demonstrate that the duration that \emph{MemX} triggers the world camera and records moments of interest accounts for a small percentage of the total usage time. \textcolor{black}{
Furthermore, in \emph{MemX}, even though the eye tracking process is always on, this stage has high energy efficiency, thanks to the low data rate of the eye camera and energy-efficient TVA network architecture design. The always-on eye-tracking process is approximately 51.98x more energy efficient than the high-resolution video capture and recording pipeline. In addition, the light-weight fusion network is approximately 43.75x more energy efficient than the VIS pipeline~\cite{zhao2021reinforcementlearningbased,yang2019video}. 
Overall, the pilot study demonstrates that, compared with the record-everything baseline, \emph{MemX} effectively improves system energy efficiency by 86.36\% on average.}
Based on the pilot studies, we estimate that, equipped with \unit[0.36]{Wh} battery (similar to Spectacles v2), \emph{MemX} is able to support 8 hours of continuous operation after fully charged, which can meet typical daily usage requirement without frequent charging.

\subsection{Discussion}

After the pilot studies, we have conducted a questionnaire involving the 11 users who participated in the pilot studies to explore 
potential personal usage scenarios. All 11 participants agree that human visual attention may serve as an easy-to-use information filter and event detection mechanism for visual content gathering. Based on their feedback, the top-3 popular potential usage scenarios include sightseeing, lifelogging, and sports logging. Some of their comments are quoted below. 
\emph{``MemX is pretty cool when I do sightseeing. With MemX, I can record wonderful scenery and interesting encounters effortlessly.''}
\emph{``I can hand-freely record memorable moments, e.g., social gathering,  during my daily life. ''}
\emph{``MemX is perfect for sports, such as cycling, probably a better choice than GoPro.''} 

Besides personal usage cases, we would also like to explore other domain segments, e.g., industry, education, and gaming, 
to support on-site visual information gathering and remote communication and interaction. In particular, \emph{MemX}, as a convenient 
human-computer interface method, can be integrated with the fast-growing AR and VR technologies. 
Furthermore, we are in the process of developing
a video-editing software to automatically create high-quality video journals, e.g., vlogs, using the visual moments captured by \emph{MemX}. 
Our goal is to enable a complete personal interest aware visual moment auto-capture and content creation framework, 
with the end goal of fulfilling the long-awaited vision of a personalized visual Memex.

However, we have also identified several 
limitations of the current version of \emph{MemX} in terms
of video quality. In particular, motion compensation is a must-have feature for many eyewear usage scenarios. Furthermore, 
attention-aware smart glasses may introduce privacy concerns. First, using \emph{MemX}, users can conduct scene
recording in a more \textcolor{black}{discrete} fashion. Second, content captured by \emph{MemX} discloses the user's personal 
interest. \Note{***RD: One might potentially introduce additional attention events into the recorded data, thereby making it difficult to reliably draw precise conclusions about user attention from the data. However, this would negatively influence functionality and energy consumption.} Our future work will focus on further improving the video quality, and more importantly, address the 
privacy concerns introduced by attention-aware personalized moment auto-capture devices.

\section{Conclusions}
\label{sctn::cnclusn}
This work aims to realize the decades-long vision of the personalized visual Memex, emphasizing the importance of content capture which must reflect personal interest to stay relevant and valuable. We have developed~\emph{MemX}, a biologically-inspired attention-aware eyewear system to enable auto-capture of personalized attentive visual content, and record moments of personal interest in the form of compact video snippets. \emph{MemX} is equipped with a new temporal visual attention (TVA) network, which unifies eye-tracking and video \textcolor{black}{analysis} to enable accurate and computation-efficient human visual attention tracking and salient visual content analysis. \emph{MemX} is evaluated using the YouTube-VIS dataset and 30 participants. Our results show that,  compared with the \textcolor{black}{eye-tracking-alone} method, \emph{MemX} significantly improves the attention tracking accuracy, while maintaining high system energy efficiency. In addition, we have conducted 11 in-field pilot studies with different potential usage scenarios, which demonstrate the feasibility and potential benefits of \emph{MemX}. We envision that \emph{MemX} can potentially benefit a wide range of personal visual content capture scenarios, such as sightseeing, lifelogging, travel experience recording, and event abstraction. 

\section*{Acknowledgments}
\textcolor{black}{This work was supported in part by the National Natural Science Foundation of China under Grant No. 62090025 and 61932007 and in part by the National Science Foundation of the United States under grant CNS-2008151.}

\bibliographystyle{ACM-Reference-Format}
\bibliography{reference}
\end{document}